\documentclass[journal]{IEEEtran}

\usepackage{xcolor,soul,framed} 

\colorlet{shadecolor}{yellow}
\usepackage[pdftex]{graphicx}
\graphicspath{{../pdf/}{../jpeg/}}
\DeclareGraphicsExtensions{.pdf,.jpeg,.png}

\usepackage[cmex10]{amsmath}
\usepackage{array}
\usepackage{mdwmath}
\usepackage{mdwtab}
\usepackage{eqparbox}
\usepackage{url}
\usepackage{cite}
\usepackage{amsmath,amssymb,amsfonts}
\usepackage{algorithmic}
\usepackage{textcomp}
\usepackage{lettrine}
\usepackage{graphicx}
\usepackage{caption}
\usepackage{subcaption}
\usepackage[export]{adjustbox}
\usepackage[utf8x]{inputenc} 
\newcommand{\subparagraph}{}
\usepackage{makecell}
\let\labelindent\relax
\usepackage{enumitem}
\usepackage{multicol,booktabs,tabularx}
\usepackage{multirow}
\usepackage{color}
\usepackage{xspace}
\usepackage{acronym}
\usepackage{graphicx}
\usepackage{amssymb}
\usepackage{pifont}
\newcommand{\cxmark}{\ding{51}}
\newcommand{\cross}{\ding{55}}

\newlist{tabitemize}{itemize}{1}
\newlist{soloitemize}{itemize}{1}
\setlist[tabitemize,soloitemize]{
  nosep, nolistsep,
  topsep=6pt,
  align=left,
  left=0pt,
  label=$\bullet$,
}

\acrodef{DL}{Deep Learning}
\acrodef{NN}{Neural Network}
\acrodef{VO}{Visual Odometry}
\acrodef{RoI}{Region of Interest}
\acrodef{RTK}{Real-Time Kinematic}
\acrodef{SA}{Situational Awareness}
\acrodef{DVS}{Dynamic Vision Sensor}
\acrodef{EKF}{Extended Kalman Filter}
\acrodef{AI}{Artificial Intelligence}
\acrodef{S-Graph}{Situational Graph}
\acrodef{DSG}{Dynamic Scene Graph}
\acrodef{UKF}{Unscented Kalman Filter}
\acrodef{SVM}{Support Vector Machines}
\acrodef{VIO}{Visual-Inertial Odometry}
\acrodef{MCL}{Monte Carlo Localization}
\acrodef{MHE}{Moving Horizon Estimation}
\acrodef{GPS}{Global Positioning System}
\acrodef{IMU}{Inertial Measurement Unit}
\acrodef{RF}{Radio Frequency}
\acrodef{SURF}{Speeded-Up Robust Features}
\acrodef{CNN}{Convolutional Neural Network}
\acrodef{RNN}{Recurrent Neural Network}
\acrodef{GNN}{Graph Neural Network}
\acrodef{LIDAR}{Light Detection and Ranging}
\acrodef{HOG}{Histogram of Oriented Gradients}
\acrodef{SIFT}{Scale-Invariant Feature Transform}
\acrodef{GNSS}{Global Navigation Satellite Systems}
\acrodef{ATIS}{Asynchronous Time-based Image Sensor}
\acrodef{SLAM}{Simultaneous Localization and Mapping}
\acrodef{DAVIS}{Dynamic and Active-pixel Vision Sensor}
\acrodef{AV}{Autonomous Vehicle}
\acrodef{SDF}{Signed Distance Field}
\acrodef{TSDF}{Truncated Signed Distance Field}
\acrodef{ESDF}{Euclidean Signed Distance Field}

\def\BibTeX{{\rm B\kern-.05em{\sc i\kern-.025em b}\kern-.08em
    T\kern-.1667em\lower.7ex\hbox{E}\kern-.125emX}}
\newcommand{\wrt}{w.r.t. }

\newcommand{\eg}{\textit{e.g., }}
\newcommand{\ie}{\textit{i.e., }}

\begin{document}
\bstctlcite{}
  \title{From SLAM to Situational Awareness: Challenges and Survey}
  \author{Hriday Bavle, Jose Luis Sanchez-Lopez, Claudio Cimarelli, Ali Tourani and Holger Voos
  \thanks{This work was partially funded by the Fonds National de la Recherche of Luxembourg (FNR), under the projects C19/IS/13713801/5G-Sky, and by a partnership between the Interdisciplinary Center for Security Reliability and Trust (SnT) of the University of Luxembourg and Stugalux Construction S.A.  
  For the purpose of Open Access, the author has applied a CC BY public copyright license to any Author Accepted Manuscript version arising from this submission
  
  Hriday Bavle, Jose Luis Sanchez-Lopez, Claudio Cimarelli, Ali Tourani, and  Holger Voos are with the Automation and Robotics Research Group, Interdisciplinary Centre for Security, Reliability and Trust, University of Luxembourg. \{hriday.bavle, joseluis.sanchezlopez, claudio.cimarelli, ali.tourani, holger.voos\}@uni.lu
  
  Holger Voos is also associated with the Faculty of Science, Technology and Medicine, University of Luxembourg, Luxembourg    
    
  }}  
    
\maketitle

\begin{abstract}
The capability of a mobile robot to efficiently and safely perform complex missions is limited by its knowledge of the environment, namely the \textit{situation}.
Advanced reasoning, decision-making, and execution skills enable an intelligent agent to act autonomously in unknown environments.
\ac{SA} is a fundamental capability of humans that has been deeply studied in various fields, such as psychology, military, aerospace, and education.
Nevertheless, it has yet to be considered in robotics, which has focused on single compartmentalized concepts such as sensing, spatial perception, sensor fusion, state estimation, and \ac{SLAM}. 
Hence, the present research aims to connect the broad multidisciplinary existing knowledge to pave the way for a complete \ac{SA} system for mobile robotics that we deem paramount for autonomy. To this aim, we define the principal components to structure a robotic \ac{SA} and their area of competence.
Accordingly, this paper investigates each aspect of \ac{SA}, surveying the state-of-the-art robotics algorithms that cover them, and discusses their current limitations.
Remarkably, essential aspects of \ac{SA} are still immature since the current algorithmic development restricts their performance to only specific environments.
Nevertheless, \ac{AI}, particularly \ac{DL}, has brought new methods to bridge the gap that maintains these fields apart from the deployment to real-world scenarios.
Furthermore, an opportunity has been discovered to interconnect the vastly fragmented space of robotic comprehension algorithms through the mechanism of \textit{\ac{S-Graph}}, a generalization of the well-known scene graph.
Therefore, we finally shape our vision for the future of robotic \acl{SA} by discussing interesting recent research directions.
\end{abstract}

\section{Introduction}
\label{sec:intro}

\begin{figure}[t]
    \centering
    \includegraphics[width=0.5\textwidth]{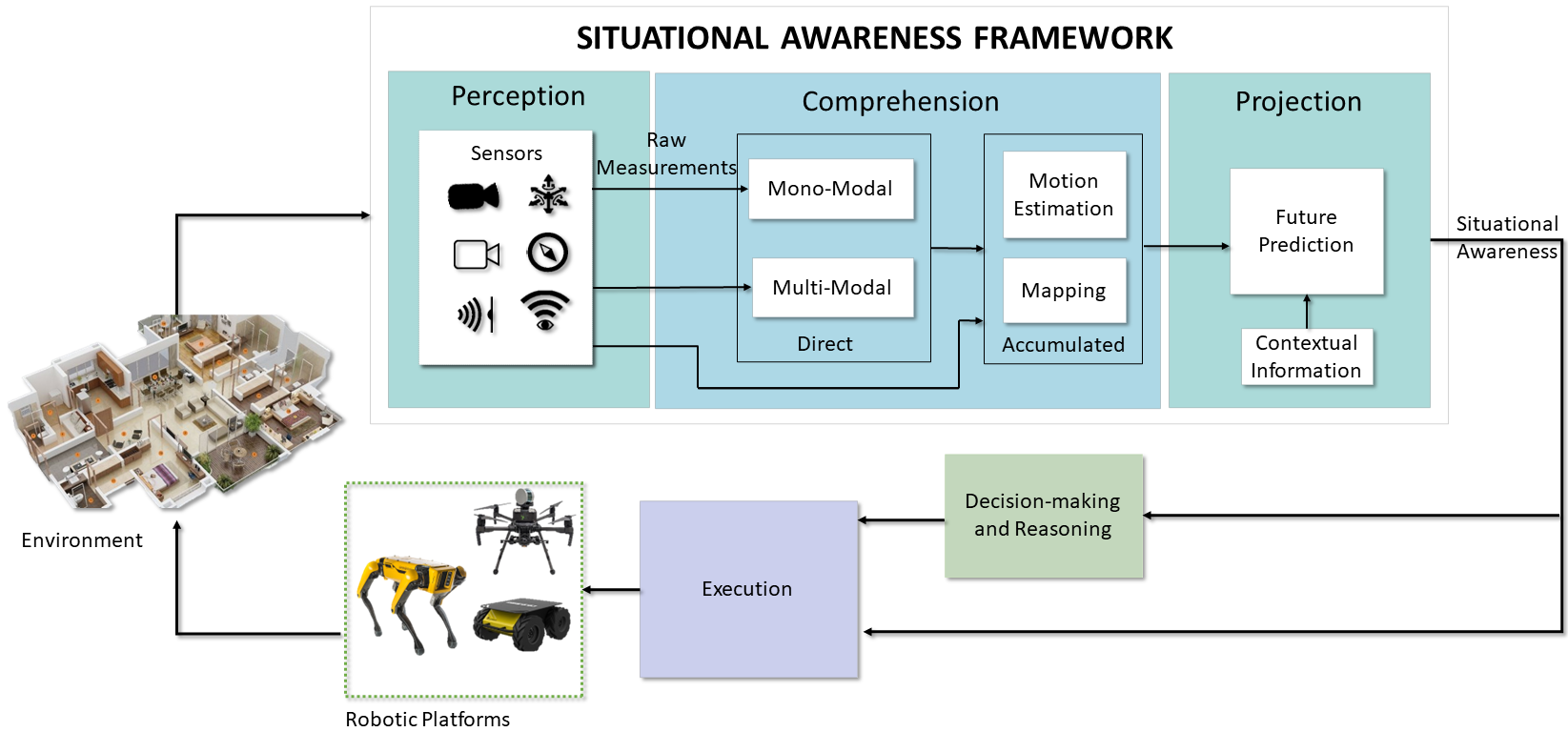}
    \caption{The proposed \acl{SA} system architecture for autonomous mobile robots. We break it into its principal components, namely Perception, Comprehension, and Projection, and show how they are connected.} 
    \label{fig:architecture}
\end{figure}

\begin{figure*}[t]
    \centering
    \includegraphics[width=0.7\textwidth]{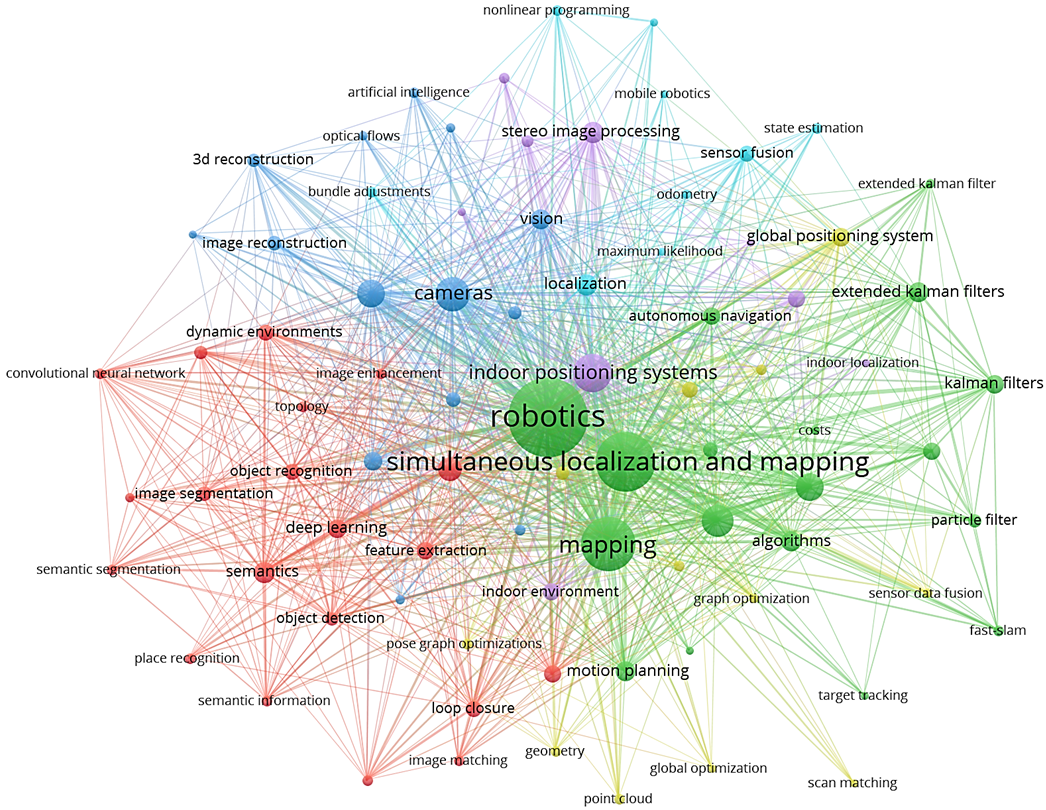}
    \caption{Scopus database since 2015 covering the research in \textit{Robotics} and \textit{\ac{SLAM}}. All the works have focused on independent research areas which could be efficiently encompassed in one field of \acl{SA} for robots.}
    \label{fig:robotics_database}
\end{figure*}

The robotics industry is experiencing exponential growth, embarking on newer technological advancements and applications.
Mobile robots have gained interest from a commercial perspective due to their capabilities to replace or aid humans in repetitive or dangerous tasks~\cite{tzafestas2018mobile}.
Already, a wide range of industrial and civil-related applications employ mobile robots~\cite{Dzedzickis2021}.
For example, industrial machines and underground mines inspections, surveillance and road-traffic monitoring, civil engineering, agriculture, healthcare, search, and rescue interventions in extreme environments, \eg natural disasters, for exploration and logistics~\cite{Siciliano2008}.

Mobile robots can be controlled in manual teleoperation or semi-autonomous mode with constant human intervention in the loop.
Instead, in fully autonomous mode, a robot performs an entire mission based on its understanding of the environment given only a few commands~\cite{siegwart2011introduction}.
Remarkably, autonomy reduces costs and risks while increasing productivity and is the goal of current research to solve the main challenges that it raises~\cite{Wong2018}.
Unlike the industrial scenario, where autonomous agents can act in a controlled environment, mobile robots operate in the dynamic, unstructured, and cluttered world domain with little or zero previous knowledge of the scene structure.

Up to now, the robotics community has focused chiefly on research areas such as sensing, perception, sensor fusion, data analysis, state estimation, localization and mapping, \ie~\acf{SLAM}, and \acf{AI} applied to various image processing problems, in a compartmentalized manner.
Fig.~\ref{fig:robotics_database} shows the mentioned targets data obtained from \textit{Scopus} abstract and citation database. However, autonomous behavior entails understanding the situation encompassing multiple interdisciplinary aspects of robotics, from perception, control, and planning to human-robot interaction.
Although \ac{SA}~\cite{situational_salas} is a holistic concept widely studied in fields like psychology, military, and aerospace, it has been barely considered in robotics.
Notably, Endsley~\cite{endsley} formally defined \ac{SA} in the 90s as "\textit{the perception of the elements in the environment within a volume of time and space, the comprehension of their meaning and the projection of their status shortly}," which remains valid till date~\cite{Munir2022}.
Hence, we turn this definition into the perspective of mobile robotics to derive a unified field of research that garners all the aspects required by an autonomous system.

Therefore, a robot's \acf{SA} system must continuously acquire new observations of the surroundings, understand its essential elements and make complex reasoning, and project the world state into a possible future outcome to make decisions and execute actions that would let it accomplish its goals.
Accordingly, we depict in Fig.~\ref{fig:architecture} a general architecture of \ac{SA} that divides the specific competence areas into three layers, with an increasing level of intelligence.
Thus, we raise the following research question:

\begin{itemize}
    \item \textit{What are the components of a robot's situational awareness system?}
\end{itemize}

To answer the mentioned question, we characterize \ac{SA} by its three main parts, for which we propose a description that delimits their scope and defines their purpose:

The \textbf{perception of the situation} consists of the acquisition of exteroceptive, \ie of the surroundings, such as visual light intensity or distance, and proprioceptive, \ie of the internal values of the robot, such as velocity or temperature, and information of the situation.
Sensors provide raw measurements that must be transformed to acquire actual knowledge or may directly inform the robot about its state with little processing.
For instance, active range sensors provide distance to objects by well-defined models.
On the contrary, the pixel intensity values of a camera, other than being distorted by unknown parameters, depend on complex algorithms to extract meaningful depth, which is still ongoing research.
Considering this, multiple sensor modalities are essential to perceive complementary details of the situation, \eg acceleration of the robot and the metric scale, the visible light intensity and its changes, and compensate for low performance in different conditions, \eg low-light, light-transparent materials, or fast motion.
Hence, the perception includes the array of sensors that give each robot specific attributes and those algorithms that augment the amount of information at the disposal of successive layers.

The \textbf{comprehension of the situation} extends from understanding the current perception, considering the possible semantic relationships, to build a short-term understanding using perceptual observation at a given time instant referred to as \textit{direct situational comprehension} or long-term model that includes the knowledge acquired in the past namely \textit{accumulated situational comprehension}.
Multiple abstract relationships can be created to link concepts in a structured model of the situation, such as geometric (\eg shape of the objects), semantic (\eg type and functionality of the objects), topological (\eg order in the space), ontological (\eg hierarchy of commonsense concepts), dynamic (\eg motion between the objects), or stochastic (\eg to include uncertainty information).
In addition, the comprehension of the situation is affected by mechanisms, such as attention, that are controlled by the decision-making and control processes (\eg looking for a particular object in a room vs getting a global overview of a room). 

The \textbf{projection of the situation} into the future is essential for decision-making processes, and a higher level of comprehension facilitates this ability. A more profound understanding of the environmental context, which includes information such as the robot's position, velocity, pose, and any static or dynamic obstacles in the surrounding area, can lead to a more accurate projection model. The projection process involves forecasting the future states of both the ego-agent and external agents to predict behaviors and interactions, enabling the robot to adapt its actions to achieve its goals effectively.

The rest of this paper aims to delve into those research questions naturally developed as a consequence of the exposed \ac{SA} topic:

\begin{itemize}
    \item \textit{What has been achieved so far, and what challenges remain?}
    \item \textit{What could be the future direction of \acl{SA}?}
\end{itemize}

Thus, by reviewing the current state-of-the-art methods for mobile robots that may fall into perception, comprehension, and projection, we aim to study the broad field of \acl{SA} as one and understand the advancement and limitations of its components.
Then, we discuss in which direction we envision the research will address the remaining challenges and bridge the gap that divides robots from mature, intelligent autonomous systems. We summarize the main contributions of this paper as:

\begin{itemize}
    \item Comprehensive review of the state-of-the-art approaches: We conduct a thorough analysis of the latest research related to enhancing situational awareness for mobile robotic platforms, covering computer vision, deep learning, and SLAM techniques. 
    \item Identification and analysis of the challenges: We classify and discuss the reviewed approaches according to the proposed definition of situational awareness for mobile robots and highlight their current limitations for achieving complete autonomy in mobile robotics.
    \item Proposals for future research directions: We provide valuable insights and suggestions for future research directions and open problems that need to be addressed to develop efficient and effective situational awareness systems for mobile robotic platforms.
\end{itemize}
\section{Situational Perception}
\label{sec:perception}

The continuous technological advances regarding chip developments have made many sensors suited for use onboard mobile robots~\cite{rubio2019review}, as they come with a small form factor and possibly low power consumption.
The primary sensor suite of the average robot can count on a wide array of devices, such as \acp{IMU}, magnetometers, barometers, and \ac{GNSS} receivers, \eg for the common \ac{GPS} satellite constellation.
Sensors such as \acp{IMU}, which can measure the attitude, angular velocities, and linear accelerations, are cheap and lightweight, making them ideal for running onboard any robotic platform.
Though the performance of these sensors can degrade over time due to the accumulation of errors coming from white Gaussian noise~\cite{imu_noise_modelling}.
Magnetometers are generally integrated within an \ac{IMU} sensor, measuring the accurate heading of the robotics platform relative to the earth's magnetic field.
The sensor measurements from a magnetometer can be corrupted in environments with constant magnetic fields interfering with the earth's magnetic field.
While barometers estimate the altitude changes through measured pressure changes, they suffer from bias and random noise in measurements in indoor environments due to ground/ceiling effects~\cite{barometers_as_altimeters}.
\ac{GNSS} receivers, as well as their higher-precision variants, such as \ac{RTK} or differential \ac{GNSS}, provides reliable position measurements in a global frame of reference relative to the earth.
However, these sensors can work only in uncluttered outdoor environments with multiple satellites connected or within direct line of sight with the \ac{RTK} base station~\cite{rtk_gps_errors}.

\begin{table*}[!h] 
\aboverulesep = 0pt
\belowrulesep = 0pt
\renewcommand{\arraystretch}{2.0}
\scriptsize
\setlength{\tabcolsep}{2pt}
\caption{Different types of sensors utilized onboard the mobile robots for situational perception.}.
\begin{tabular}{>{\raggedright\arraybackslash}p{0.12\linewidth} | >{\raggedright\arraybackslash}p{0.12\linewidth} | >{\raggedright\arraybackslash}p{0.17\linewidth} | >{\raggedright\arraybackslash}p{0.14\linewidth} | >{\raggedright\arraybackslash}p{0.26\linewidth} | >{\raggedright\arraybackslash}p{0.11\linewidth}}
\toprule
  Classification & Sensor  &  Measurement & Mobile Robotic Platforms & Limitations &  Examples \\
\midrule   
 \multirow{10}{*}{Proprioceptive} & IMU & \begin{soloitemize}
     \item Velocities 
    \item Accelerations 
    \item Yaw angle (\textbackslash w magnetometer) 
    \end{soloitemize}  & Indoor/Outdoor Robots & \begin{soloitemize} 
    \item Bias \item Gaussian Noise 
       \item Drift rapidly
       \end{soloitemize}  & MPU-6050 \\ \cmidrule{2-6}
   & GPS & \begin{soloitemize}
       \item Absolute Position
   \end{soloitemize} & Outdoor Robots & \begin{soloitemize}
       \item Unreliable measurements in cluttered environments.
   \end{soloitemize}  & u-blox NEO-M8N \\ \cmidrule{2-6}
   & Barometer & \begin{soloitemize}
       \item Altitude from Atmospheric Pressure
   \end{soloitemize} & {Indoor/Outdoor Aerial Robots} & \begin{soloitemize}
       \item Bias 
       \item Gaussian Noise
   \end{soloitemize} {} & {Bosch BMP280} \\ \cmidrule{2-6}
   & Robot Encoders & \begin{soloitemize}
       \item Relative Position 
       \item Velocity
   \end{soloitemize} & {Indoor/Outdoor Ground Robots} & \begin{soloitemize}
       \item Slippage
       \item Error Accumulation
   \end{soloitemize} & {US Digital E4P}  \\ \cmidrule{2-6}
   & RF Receiver & \begin{soloitemize}
       \item Absolute Position 
   \end{soloitemize} & Indoor/Outdoor Robots & \begin{soloitemize}
       \item Prone to Interference 
       \item Limited Range
   \end{soloitemize} &{DecaWave DWM1000} \\
\midrule   
\multirow{18}*{Extereoceptive} & RGB Camera & \begin{soloitemize}
    \item Visible Light 
\end{soloitemize} & Indoor/Outdoor Robots & 
    \begin{soloitemize}
        \item Motion Blur 
        \item Degradation in Poor Light Conditions
    \end{soloitemize} & IDS uEye LE \\ \cmidrule{2-6}
   & RGB-D Camera &  \begin{soloitemize}
       \item Visible Light 
       \item Depth from IR Structured Light
   \end{soloitemize} & Indoor/Outdoor Robots & \begin{soloitemize}
       \item Limited and Noisy Range
       \item Errors in Reflective/Transparent Surfaces
   \end{soloitemize}
   & {Intel Realsense D435} \\ \cmidrule{2-6}
   & IR Camera & \begin{soloitemize}
       \item Infrared Radiation 
   \end{soloitemize} & Indoor/Outdoor Robots & \begin{soloitemize}
       \item Limited Information 
       \item Prone to Atmospheric Interference
       \item Infrared Cannot Pass Through Glass or Water
   \end{soloitemize}& {FLIR Lepton}\\ \cmidrule{2-6}
   & Event Camera &  \begin{soloitemize}
       \item Brightness Log-Intensity Changes
   \end{soloitemize} & Indoor/Outdoor Robots & \begin{soloitemize}
       \item Requires Motion of Camera or Objects
       \item Absolute Brightness Not Measured Directly
       \item Not Easy to Purchase
   \end{soloitemize} & DAVIS 346 or SONY IMX636ES \\ \cmidrule{2-6}
   & LIDAR & \begin{soloitemize}
       \item Metric Distances and Angle of Scene Points
   \end{soloitemize} & Indoor/Outdoor Robots  & \begin{soloitemize}
       \item Prone to Atmospheric Interference
       \item Degradation in Reflective and Transparent Surfaces
   \end{soloitemize}
    & {Velodyne VLP-16} \\ \cmidrule{2-6}
   & MmWave FMCW RADAR & \begin{soloitemize}
       \item Metric Distances and Angle of Scene Points
       \item Objects speed
   \end{soloitemize} & Indoor/Outdoor Robots & \begin{soloitemize}
     \item Limited Range and Field of View
       \item Possible Low Angular and Distance Resolution
       \item Multipath Propagation Effect and Ghost Targets
   \end{soloitemize}
    &  AWR6843AOP \\ 
\bottomrule 
\end{tabular}  \label{tab:sensor_summary}
\end{table*}

The adoption of cameras as exteroceptive sensors in robotics has become increasingly prevalent due to their ability to provide a vast range of information in a compact and cost-effective manner~\cite{tourani2022visual}.
In particular, RGB cameras, including monocular cameras, have been widely used in robotics. 
Additionally, cameras with depth information, such as stereo or RGB-D cameras, have emerged as a dominant sensor type in robotics. 
This is because they enable the extraction of a broad range of knowledge about the surrounding environment and are easily interpretable by humans.
As such, the use of standard cameras is expected to continue playing a crucial role in developing advanced robotic systems.
These cameras suffer from the disadvantages of motion blur in the presence of rapid motion of the robots, and the perceived quality can be degraded in changing lighting conditions.

In addition to RGB and depth cameras, thermal and infrared (IR) cameras are frequently utilized in robotics, exceptionally when visibility is limited due to nighttime or adverse weather conditions. 
These sensors can provide valuable information not detectable by human eyes or traditional cameras, such as heat signatures and thermal patterns. 
By incorporating thermal and IR cameras into the sensor suite, mobile robots can detect and track animated targets by following heat signatures, navigating low-visibility environments, and operating in a broader range of conditions. 
Thus, these specialized sensors can significantly enhance robotic systems' situational awareness and overall performance.

Event cameras, also known as neuromorphic sensors, such as the \ac{DVS}~\cite{dvs}, overcome these limitations by encoding pixel intensity changes rather than an absolute brightness value and providing very high dynamic ranges as well as no motion blur during rapid motions.
However, due to the asynchronous nature of the event camera, measurements of the situations are only provided in case of variations in the perceived scene brightness that are often caused by the motion of the sensor itself.
Hence, they can measure a sparse set of points, usually in correspondence with edges.
To perceive a complete dense representation of the environment, such sensors are typically combined with traditional pixels, as in the case of the \ac{DAVIS}~\cite{davis} or \ac{ATIS}~\cite{atis} cameras.
Still, algorithms have also been proposed to reconstruct images by integrating events over time~\cite{event-camera-survey}.

Ranging sensors, such as small factor solid-state \acp{LIDAR} or ultrasound sensors, are the second most dominant group of employed exteroceptive sensors.
1D \acp{LIDAR} and ultrasound sensors are used mainly in aerial robots to measure their flight altitude but only measure limited information about their environments.
2D and 3D \acp{LIDAR} accurately perceive the surroundings in 360\textdegree~and the newer technological advancements have reduced their size and weight.
However, utilizing these sensors onboard small-sized robotic platforms is still not feasible, and the high acquisition cost hampers the adoption of this sensor to the broad commercial market.
Even for autonomous cars, a pure-vision system, which may include event cameras, is often more desirable from an economic perspective.

Millimeter-wave (mmWave) radar sensors are an alternative to LIDAR for range measurements in robotics~\cite{venon2022millimeter}. 
While they have a lower angular resolution and limited range than LIDAR, they offer a smaller form factor and a lower cost. 
MmWave radars transmit radio waves and detect their reflections off objects in their field of view.
This lets them determine detected objects' range, velocity, and angle. 
Moreover, mmWave radars can detect transparent surfaces that are challenging to see with other types of sensors, \eg \ac{LIDAR}. 
As such, they have become an attractive option for robotic applications where cost, form factor, and detection of transparent surfaces are crucial.

\ac{RF} signal is another technology based on signal processing that allows a robot to infer its global position by estimating its distance with one or multiple base stations. Differently from \ac{GPS}, \ac{RF} may be able to provide positioning information also in indoor environments, even though range measurements require it to be fused with other sources of motion estimation, \eg from \ac{IMU}. Contrary to mmWave radars, \ac{RF}-based localization or even mapping is far less precise, but newer technology such as 5g promises superior performances. However, a drawback of these approaches is that they require synchronization between the antenna and the receiver for computing a time-of-arrival and possibly line-of-sight communication when only one antenna is available. \cite{Kabiri2023} provide an exhaustive review of \ac{RF}-based localization methods and give an outlook on current challenges and future research directions.

\begin{figure*}[t]
    \centering
    \includegraphics[width=1.0\textwidth]{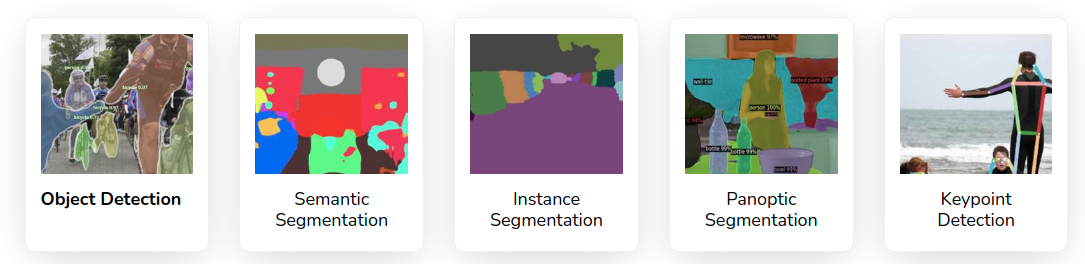}
    \label{fig:different_detections}
    \caption{Deep learning-based computer vision algorithms for mono-modal scene understanding. Copyright to~\cite{paperwithcode}.}
\end{figure*}

Table.~\ref{tab:sensor_summary} summarizes the different sensors used onboard mobile robots with their individual limitations. Due to the multi-faceted characteristics of the available sensors, relying on a mono-modal perception does not guarantee a safe robot deployment in real-world settings.
Consequently, multi-modal perception is often preferred at the cost of more complex solutions to fuse and time-synchronize the measurements from multiple sensors properly.
Nevertheless, it is essential to perceive complementary information of the situation and to build a complete state of the autonomous agent, \eg acceleration of the robot, the visual light intensity of the environment, its global position, the distance with obstacles, and compensate for low performance in different conditions, \eg dark rooms or low-light, transparent materials or non-Lambertian surfaces.


The traditional approach to designing robots involves tailoring their sensor selection and configuration to the specific task they are intended to perform. However, this approach may not be sufficient for a human-like perception capable of adapting to diverse external situations. To overcome this limitation, it is necessary to equip robots with a standard, versatile sensor suite that can provide detailed and accurate information about their surroundings and dynamic elements. This sensor suite, coupled with advanced processing algorithms (explained in Section.~\ref{subsec:scene_understanding} and Section.~\ref{subsec:long_term}), can enable robots to perceive their environment like humans, irrespective of the application they are designed for.



\section{Direct Situational Comprehension}
\label{subsec:scene_understanding}

Some research works focus on transforming the complex raw measurements provided by the sensors into more tractable information with different levels of abstraction, \ie feature extraction for an accurate scene understanding, without building a complex long-term model of the situation.
Direct situational comprehension based on the sensor modalities can be divided into two main categories, as described below:

\subsection{Mono-Modal}
\label{subsec:monomodal_scene_understanding}

These algorithms utilize a single sensor source to extract useful environmental information. The two major sensor modalities used in robotics are \textit{vision-based sensors} and \textit{range-based sensors} for the rich and plentiful amount of information in their scene observations.

Vision-based comprehension started with the early works of Viola and Jones~\cite{viola_and_jones} presenting an object-based detector for face detection using \textit{Haar-like features} and \textit{Adaboost feature classification}.
Following works for visual detection and classification tasks such as~\cite{obj-det-using-sift,img-clas-using-sift,hog_based_human_det,mono_pedestrian_detection,vision_based_bicycle} utilized well-known image features, \eg \ac{SIFT}~\cite{sift}, \ac{SURF}~\cite{surf}, \ac{HOG}~\cite{hog}, along with \ac{SVM}-based classifiers~\cite{svm}.
The mentioned methods focused on extracting only a handful of useful information from the environment, such as pedestrians, cars, and bicycles, showing degraded performance in difficult lighting conditions and occlusions.

With the establishment of \ac{DL} in computer vision and robotic image processing, recent algorithms in the literature robustly extract the scene information  utilizing \acp{CNN} in the presence of different lighting conditions and occlusions. 
In computer vision, different types of \ac{DL}-based methods exist based on the type of extracted scene information.
Algorithms such as \textit{Mask-RCNN}~\cite{mask_rcnn}, \textit{RetinaNet}~\cite{retinanet}, \textit{TensorMask}~\cite{tensormask}, \textit{TridentNet}~\cite{tridentnet}, and \textit{Yolo}~\cite{yolov4} perform detection and classification of several object instances, they either provide a bounding box around the object or perform a pixel-wise segmentation for each object instances.
Other algorithms such as~\cite{semantic_jlong,encoder_decoder_chen,pointrend,fast_scnn,deeplabv3} perform dense semantic segmentation, being able to extract all relevant information from the scene.
Additionally,~\cite{panoptic_kirillov, panopticdeeplab} aim to detect and categorize all object instances in an image through panoptic segmentation, regardless of their size or position, while still maintaining a semantic understanding of the scene.
This task is particularly challenging because it requires integrating pixel-level semantics and instance-level information. 2D \textit{scene graphs} \cite{Xu_2017_CVPR, Zareian2020} could then connect the semantic elements detected by panoptic segmentation in a knowledge graph that let reason about relationships. Moreover, this knowledge graph facilitates inferring single behaviors and interactions among the participants in a scene, animate or inanimate~\cite{suhail2021energy}.  

To overcome the limitations of the visible spectrum in the absence of light, thermal infrared sensors have been researched to augment situational comprehension.
For instance, one of the earlier methods~\cite{thermal_hum_det} finds humans in nighttime images by extracting thermal shape descriptors that are then processed by \textit{Adaboost} to identify positive detection.
In contrast, newer methods~\cite{yolo_thermal, obj_det_ther_img} utilize deep \acp{CNN} on thermal images for identifying different objects in the scene, such as humans, bikes, and cars.
Though research in the field of event-based cameras for scene understanding is not yet broad, some works such as~\cite{dvs_dynamic_obj_det} present an approach for dynamic object detection and tracking using the event streams, whereas~\cite{event_cnn} present an asynchronous \ac{CNN} for detecting and classifying objects in real-time.
\textit{Ev-SegNet}~\cite{ev_segnet} is an approach that introduced one of the first semantic segmentation pipelines based on event-only information.

\begin{figure*}[h]
    \begin{subfigure}{.5\textwidth}
    \centering
    \includegraphics[width=0.95\textwidth,  height=39mm, keepaspectratio]{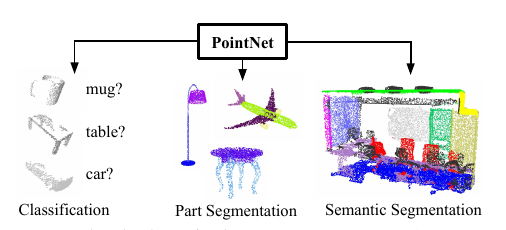}
    \label{fig:pointnet}
    \caption{}
    \end{subfigure}
    \begin{subfigure}{.5\textwidth}
    \centering
    \includegraphics[width=1.0\textwidth]{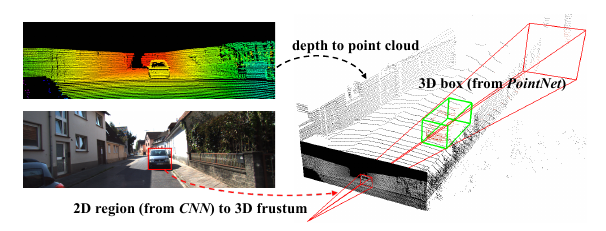}
    \label{fig:frustrumpointnet}
    \caption{}
    \end{subfigure}
    \caption{Mono-modal and multi-modal scene understanding algorithms, (a) \textit{PointNet} algorithm using only \ac{LIDAR} measurements. Copyright to~\cite{pointnet}. (b) \textit{Frustrum PointNets} algorithm combining RGB and \ac{LIDAR} measurements improving the accuracy of \textit{PointNet}. Copyright to~\cite{frustum_pointnet}.}
\end{figure*}

\begin{table*}[t] 
\aboverulesep = 0pt
\belowrulesep = 0pt
\renewcommand{\arraystretch}{2.0}
\scriptsize
\setlength{\tabcolsep}{2pt}
\caption{Summary of the algorithms for the direct situational comprehension \ac{SA} module. DL Refers to methods leveraging Deep Learning.}.
\begin{tabular}{>{\raggedright\arraybackslash}p{0.11\linewidth} | >{\raggedright\arraybackslash}p{0.10\linewidth} | >{\raggedright\arraybackslash}p{0.21\linewidth} | >
{\centering\arraybackslash}p{0.03\linewidth} | >
{\raggedright\arraybackslash}p{0.35\linewidth}  | >{\raggedright\arraybackslash}p{0.11\linewidth}}
\toprule
Modality & Sensor &  Method & DL & Limitations & References \\
\midrule 
\multirow{22}*{Mono-Modal} & \multirow{4}*{RGB} & Feature Detection & \cross & \begin{soloitemize}   
     \item Sensitive to Illumination Changes 
     \item Higher False Positives 
     \item Lower Robustness in the presence of occlusions
     \end{soloitemize}
 & \cite{viola_and_jones, obj-det-using-sift,img-clas-using-sift,hog_based_human_det,mono_pedestrian_detection,vision_based_bicycle, sift, surf, hog, svm} \\ \cmidrule{3-6}
 & & Object Detection & \cxmark & \begin{soloitemize}   
     \item Higher Computation Cost
     \item Larger Training Data
     \item Sensitive to Occlusions
     \item No Instance Segmentation
     \end{soloitemize} & \cite{yolov4} \\ \cmidrule{3-6}
  & & Semantic Segmentation & \cxmark & \begin{soloitemize}   
     \item Higher Computation Cost
     \item Larger Training Data
     \item No Instance Segmentation
     \end{soloitemize} & \cite{mask_rcnn, retinanet, tensormask, tridentnet, semantic_jlong,encoder_decoder_chen,pointrend,fast_scnn,deeplabv3} \\ \cmidrule{3-6} 
 & & Panoptic Segmentation & \cxmark & \begin{soloitemize}   
     \item Higher Computation Cost
     \item Larger Training Data
     \item Slower Inference Time
     \end{soloitemize} & \cite{panoptic_kirillov, panopticdeeplab} \\ \cmidrule{3-6} 
 &  & 2D Scene Graphs & \cxmark & \begin{soloitemize}   
     \item Limited to 2D Spatial Information
     \item Limited Temporal Information
     \end{soloitemize} & \cite{Xu_2017_CVPR, Zareian2020, suhail2021energy} \\  \cmidrule{2-6}   
& \multirow{2}*{Thermal} & Object Detection & \cross & \begin{soloitemize}  
     \item Limited Applicability
     \item Higher False Positives
     \end{soloitemize} & \cite{thermal_hum_det} \\ \cmidrule{3-6}
 & & Object Detection & \cxmark & \begin{soloitemize}   
     \item Limited Applicability
     \item Limited Datasets
     \end{soloitemize} & \cite{yolo_thermal, obj_det_ther_img} \\ \cmidrule{2-6}
 & {Event} & Object Detection & \cxmark & \begin{soloitemize}
     \item Trained over Limited Data
     \item Limited Validation in Presence of Occlusions
 \end{soloitemize} & \cite{event_cnn, ev_segnet} \\ \cmidrule{3-6}  
 & & Semantic Segmentation & \cxmark & \begin{soloitemize}
     \item Trained over Limited Data
     \item Limited to only Few Semantic Objects
 \end{soloitemize} & \cite{ev_segnet} \\ \cmidrule{2-6}  
 & \multirow{2}*{LIDAR} & Object Detection & \cross & \begin{soloitemize}
     \item Detection of Fewer Semantic Entities
     \item Lower Robustness in Presence of Outliers and Occulusions
 \end{soloitemize} &  \cite{contour_based_object_detection, 3D_lidar_obj_det, 3D_lidar_obj_terr_clas} \\ \cmidrule{3-6}
 & & Semantic Segmentation & \cxmark & \begin{soloitemize}
     \item Limited to Fewer Semantic Entities
     \item Higher Computational Cost
     \item Lower Accuracy in Indoor Environments
 \end{soloitemize} & \cite{rangenet++, learning_3D_in_2D, squeeze_seg, squeeze_seg2, pointnet, pointnet_plus_plus, tangent_convolutions, dops_2020, randlanet} \\   \midrule
\multirow{12}*{Multi-Modal} & \multirow{2}*{RGB+Depth} & Object Detection & \cross & \begin{soloitemize}
     \item Higher False Positives and Negatives
     \item Limited Range
 \end{soloitemize} & \cite{on-board-object-det, holistic_scene_und} \\ \cmidrule{3-6}
 &  & Object Detection & \cxmark & \begin{soloitemize}
     \item Limited Range and limited to Low Range Applications
     \item High Computation Cost 
     \item Lower Inference Time
     \item Lack of Generalizability over non-trained Semantic Entities
 \end{soloitemize} & \cite{RGBD_obj_det, posecnn, densefusion,normalized_coordinate_space,multiview} \\ \cmidrule{2-6}
 & \multirow{2}*{\makecell[l]{RGB + \\Thermal}} & Semantic Segmentation & \cxmark & \begin{soloitemize}
     \item Limited Real World Testing
     \item Mostly Limited to Outdoor Environments 
     \item Lower Object Detection Accuracy
 \end{soloitemize} & \cite{MFNet,rtfnet,PST900, FuseSeg} \\ \cmidrule{2-6}
&  \multirow{1}*{RGB+Event} & Semantic Segmentation & \cxmark & \begin{soloitemize}
    \item Tested only on Outdoor Datasets
\end{soloitemize} & \cite{issafe} \\ \cmidrule{2-6}
 &  \multirow{1}*{RGB+LIDAR} & Object Detection & \cxmark & \begin{soloitemize}
     \item Limited accuracy in Indoor Environments
     \item No Temporal History of Detected Objects for Efficient Tracking 
 \end{soloitemize} & \cite{frustum_pointnet, 3d_proposal_obj_det, deep_cont_fusion, multiview_chen, point_fusion} \\ 
\bottomrule
\end{tabular}
\label{tab:direct_comprehension_summary}
\end{table*}

Range-based comprehension methods with earlier works such as~\cite{contour_based_object_detection} and~\cite{3D_lidar_obj_det} present object detection for range images from 3D \ac{LIDAR} using an \ac{SVM} for object classification.
However, authors in~\cite{3D_lidar_obj_terr_clas} utilize range information to identify terrain around the robot and objects and use \acp{SVM} to classify each category.
Nowadays, deep learning is also playing a fundamental role in scene understanding using range information. 
Some techniques utilize \acp{CNN} for analyzing range measurements translated into camera frames by projecting the 3D points onto an abstract image plane. 
For example, \textit{Rangenet++}~\cite{rangenet++},~\cite{learning_3D_in_2D}, \textit{SqueezeSeg}~\cite{squeeze_seg}, and \textit{SqueezeSegv2}~\cite{squeeze_seg2} project the 3D point-cloud information onto 2D range based images for performing the scene understanding tasks.
The methods mentioned above argue that \ac{CNN}-based algorithms can be directly applied to range images without using expensive 3D convolution operators for point cloud data.
Others apply \acp{CNN} directly on the point cloud information for maximizing the preservation of spatial information. 
Approaches such as \textit{PointNet}~\cite{pointnet}, \textit{PointNet++}~\cite{pointnet_plus_plus}, \textit{TangentConvolutions}~\cite{tangent_convolutions}, \textit{DOPS}~\cite{dops_2020}, and \textit{RandLA-Net}~\cite{randlanet}, perform convolutions directly over the 3D point cloud data in order to semantically label the point cloud measurements.


\subsection{Multi-Modal}
\label{subsec:multimodal_scene_understanding}

The fusion of multiple sensors for situational comprehension allows algorithms to increase accuracy by observing and characterizing the same environment quantity but with different sensor modalities~\cite{intro-msf}.
Algorithms combining RGB and depth information have been widely researched due to the easy availability of the sensors publishing RBG-D information.
\cite{on-board-object-det} study and present the improvement of the fusion of multiple sensor modalities (RGB and depth images), multiple image cues, and multiple image viewpoints for object detection.
Whereas~\cite{holistic_scene_und} combine 2D segmentation and 3D geometry understanding methods to provide contextual information for classifying the categories of the objects and identifying the scene in which they are placed.
Several algorithms classifying and estimating the pose of objects using \acp{CNN}, such as~\cite{RGBD_obj_det}, \textit{PoseCNN}~\cite{posecnn}, \textit{DenseFusion}~\cite{densefusion},~\cite{normalized_coordinate_space}, and~\cite{multiview}, rely extensively on RBG-D information. Primarily, these methods are employed for object manipulation tasks, using either robotic manipulators fixed on static platforms or mobile robots.

\cite{context_aware_fusion} fuse RGB and thermal images from a video stream using contextual information to access the quality of each image stream to combine the information from the two sensors accurately.
Whereas methods such as \textit{MFNet}~\cite{MFNet}, \textit{RTFNet}~\cite{rtfnet}, \textit{PST900}~\cite{PST900}, and \textit{FuseSeg}~\cite{FuseSeg} combine the potential of RGB images along with thermal images using \ac{CNN} architectures for semantic segmentation of outdoor scenes, providing accurate segmentation results even in the presence of degraded lighting conditions.
\cite{ECFFNet} propose \textit{ECFFNet} to perform the fusion of RGB and thermal images at the feature level, which provides complementary information, effectively improving object detection in different lighting conditions.
\cite{rbgdt-fusion, trimodal_fusion} perform a fusion of RGB, depth, and thermal camera computing descriptors in all three image spaces and fuse them in a weighted average manner for efficient human detection.

\cite{rbgde_pipeline} fuse the information from an RGB and depth sensor with an event-based camera cascading the output of a deep \ac{NN} based on event frames with the output from a deep \ac{NN} for RBG-D frames for robust pose tracking of high speed moving objects.
\textit{ISSAFE}~\cite{issafe} is another approach that combines event-based \ac{CNN} with an RGB based \ac{CNN} using an attention mechanism to perform semantic segmentation of a scene, utilizing the event-based information to stabilize the semantic segmentation in the presence of high-speed object motions.

To improve situational comprehension using 3D point cloud data, methods have been presented that combine information extracted over RGB images with their 3D point cloud data to identify and localize the objects in the scene accurately.
\textit{Frustrum PointNets}~\cite{frustum_pointnet} performs 2D detection over RGB images which are projected to a 3D viewing frustum from which the corresponding 3D points are obtained, to which a \textit{PointNet}~\cite{pointnet} is applied for object instance segmentation and an \textit{Amodal} bounding box regression is performed.
Methods such as \textit{AVOD}~\cite{3d_proposal_obj_det} and~\cite{deep_cont_fusion} extract features from both RGB and 3D point clouds projected to bird's eye view and fuse them together to provide 3D bounding boxes for several object categories.
\textit{MV3D}~\cite{multiview_chen} extract features from RGB images and 3D point cloud data from the front view as well as birds' eye view to fuse them together in \ac{RoI}-pooling, predicting the bounding boxes as well as the object class.
\textit{PointFusion}~\cite{point_fusion} employs an RGB and 3D point cloud fusion architecture which is unseen and object-specific and can work with multiple sensors providing depth.

Table.~\ref{tab:direct_comprehension_summary} provides a summary of the presented direct comprehension methods with their key limitations while using onboard mobile robots. \textit{Direct Situational Comprehension} algorithms only provide the representation of the environment at a given time instant and mostly discard the previous information, not creating a long-term map of the environment.  
In this regard, the extracted knowledge can thus be transferred to the subsequent layer of \textit{Accumulated Situational Comprehension}.

\section{Accumulated Situational Comprehension}
\label{subsec:long_term}
A greater challenge consists of building a long-term multi-abstraction model of the situation, including past information. Even small errors not considered at a particular time instant can cause a high divergence between the state of the robot and the map estimate over time.
To simplify the explanation, we divide this section into three subsections, namely \textit{Motion Estimation}, \textit{Motion Estimation and Mapping}, and \textit{Mapping}.     

\subsection{Motion Estimation}
\label{subsubsec:sensor_fusion}

The motion estimation component is responsible for estimating the state of the robot directly using the sensor measurements from single/multiple sources and/or the inference provided by the \textit{direct situational comprehension} component (see Sect.~\ref{subsec:monomodal_scene_understanding}).
While some motion estimation algorithms only use the real-time sensor information to estimate the robot's state, others estimate the robot's state inside a pre-generated environment map.
Early methods estimated the state of the robot based on filtering-based sensor fusion techniques such as an \textit{\ac{EKF}}, \textit{\ac{UKF}}, and \textit{\ac{MCL}}.
Methods such as~\cite{mcl_dellaert} and~\cite{mcl_thrun} use an \textit{\ac{MCL}} providing a probabilistic hypothesis of the state of the robot directly using the range measurements from a range sensor.~\cite{fusion_ukf} perform a \textit{\ac{UKF}} based fusion of several sensor measurements such as gyroscopes, accelerometers, and wheel encoders to estimate the motion of the robot.
\cite{mrl_using_ekf, ekf_loc_teslic} perform \ac{EKF}-based fusion of odometry from robot wheel encoders and measurements from a pre-built map of line segments to estimate the robot state, whereas~\cite{chen_ekf} use a pre-built map of corner features.
\cite{ganganath_mobile_2012} present both \ac{UKF} and \ac{MCL} approaches for estimating the pose of the robot using wheel odometry measurements and a sparse pre-built map of visual markers detected from an RGB-D camera. In contrast,~\cite{dynamic_ultra_mr} present a similar approach using ultrasound distance measurements with respect to an ultrasonic transmitter. 

The simplified mathematical models are subject to several assumptions that limit earlier motion estimation methods.
Newer methods try to improve these limitations by providing mathematical improvements over the earlier methods and account for delayed measurements between different sensors, such as the \ac{UKF} developed by~\cite{ethzmsf} and an \ac{UKF} developed by~\cite{iekf}, which compensates for time delayed measurements in an iterative nature for quick convergence to the real state.
\cite{moore_ekf} present an \ac{EKF}/\ac{UKF} algorithm well known in the robotics community, which can take an arbitrary number of heterogeneous sensor measurements for estimation of the robot state.
\cite{wan_esekf} use an improved version of Kalman filter called the \textit{error state Kalman filter}, which uses measurements from \ac{RTK} \ac{GPS}, \ac{LIDAR} and \ac{IMU} for robust state estimation \cite{miukf} present a \textit{Multi-Innovation UKF (MI-UKF)}, which utilizes a history of innovations in the update stage to improve the accuracy of the state estimate, it fuses \ac{IMU}, encoder and \ac{GPS} data and estimates the slip error components of the robot.         

Motion estimation of robots using \ac{MHE} has also been studied in the literature where methods such as~\cite{mhe_lrs_odom} fuse wheel odometry and \ac{LIDAR} measurements using an \ac{MHE} scheme to estimate the state of the robot claiming robustness over the outliers in the \ac{LIDAR} measurements \cite{mhe_liu, mhe_dubois} study a \textit{multi-rate MHE} sensor fusion algorithm to account for sensor measurements obtained at different sampling rates \cite{mhe_osman} present a generic \ac{MHE} based sensor fusion framework for multiple sensors with different sampling rates, compensating for missed measurement, outlier rejection, and satisfying real-time requirements. 

\begin{figure}
    \centering
    \includegraphics[width=0.5\textwidth]{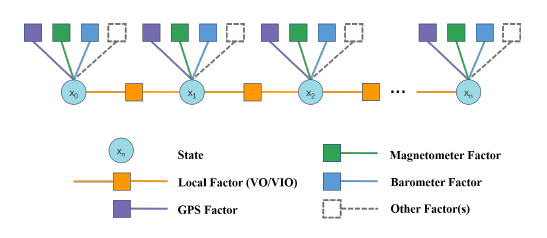}
    \caption{Localization factor graph used for estimating the robot state fusing multiple sensor measurements. Copyright to~\cite{qin2019general}.}
    \label{fig:localization}
\end{figure}

Recently, motion estimation algorithms of mobile robots using \textit{factor graph}-based approaches have also been extensively studied as they have the potential to provide higher accuracy.
Factor graphs can encode either the entire previous state of the robot or up to a fixed amount of recent states, \ie fixed-lag smoothing methods, capable of handling different sensor measurements in terms of non-linearity and varying frequencies optimally and intuitively (see Fig.~\ref{fig:localization}).
\cite{3d_pose_est_graph_based} present one of the first graph-based approaches using \textit{square root fixed-lag smoothing}~\cite{square_root_sam}, for fusing information from odometry, visual and \ac{GPS} sensors, whereas~\cite{localization_isam} present an improved fusion based on incremental smoothing approach \textit{iSAM2}~\cite{isam2} fusing \ac{IMU}, \ac{GPS} and visual measurements from a stereo-camera setup.
Methods presented in~\cite{pose_fusion_graph, merfels_sensor_fusion_2017} utilize sliding window factor graphs for estimating the robot's state by fusing several wheel odometry sources along with global pose sources \cite{gomsf} also present a sliding window factor graph fusing visual odometry information, \ac{IMU} and \ac{GPS} information to estimate the drift between the local odometry frame \wrt the global frame, instead of directly estimating the robot state \cite{qin2019general} present a generic factor graph-based framework for fusing several sensors. Each sensor serves as a factor connected with the robot's state, easily adding them to the optimization problem.
\cite{li_semi_tightly_2021} propose a novel graph-based framework for sensor fusion that combines data from a stereo visual-inertial navigation system, \ie S-VINS, and multiple \ac{GNSS} sources in a semi-tightly coupled manner. 
The S-VINS output is an initial input to the position estimate derived from the \ac{GNSS} system in challenging environments where \ac{GNSS} data is limited.
By integrating these two data sources, the framework improves the overall accuracy of the robot's global pose estimation.

The \textit{motion estimation} algorithms, as illustrated in Sect.~\ref{fig:localization}, do not simultaneously create a map of the environment, limiting their environmental knowledge, which has led to the research of simultaneous \textit{motion estimation and mapping} algorithms described in the following subsection. 

\subsection{Motion Estimation and Mapping}
\label{subsubsec:SLAM}

This section covers the approaches which estimate not only the robot motion given the sensor measurements but also the map of the environment \ie model the scene in which the robot navigates.
These approaches are commonly known as \ac{SLAM}, which is one the widely researched topics in the robotics industry~\cite{past_present_future_slam}, as it enables a robot with the capability of scene modeling without the requirement of prior maps and in applications where prior maps cannot be obtained easily.
Vision and \ac{LIDAR} sensors are the two main exteroceptive sensors used in \ac{SLAM} for map modeling~\cite{tourani2022visual}.
As in the case of \textit{motion estimation} methods, \ac{SLAM} can be performed using a single sensor modality or using information from different sensor modalities and combining it with scene information extracted from the \textit{direct situational comprehension} module (see Sect.~\ref{subsec:scene_understanding}).
\ac{SLAM} algorithms have a subset of algorithms that do not maintain the entire map of the environment and do not perform stages of \textit{loop closure} called \textit{odometry estimation algorithms}, where \ac{VO} becomes a subset of Visual-\ac{SLAM} (VSLAM) and \ac{LIDAR} odometry a subset of \ac{LIDAR} \ac{SLAM}.

\subsubsection{Filtering}

Earlier \ac{SLAM} approaches like~\cite{ekf_slam_3, ekf_slam_2, ekf_slam_1} applied \ac{EKF} for estimating the robot pose simultaneously adding/updating the landmarks observed by the robots.
However, these methods were quickly discarded as their computational complexity increased with the number of landmarks, and They did not efficiently handle non-linear measurements~\cite{consistency_ekf_slam}.
Accordingly, \textit{FastSLAM 1.0} and \textit{FastSLAM 2.0}~\cite{fastSLAM} were proposed as improvements to the \ac{EKF}-\ac{SLAM}, which combined particle filters for calculating the trajectory of the robot with individual \acp{EKF} for landmark estimation.
These techniques also suffered from the limitations of sample degeneracy when sampling the proposal distribution and the problems with particle depletion.

\begin{figure*}[ht]
    \centering
    \begin{subfigure}{.4\textwidth}
    \centering
    \includegraphics[width=1.0\textwidth]{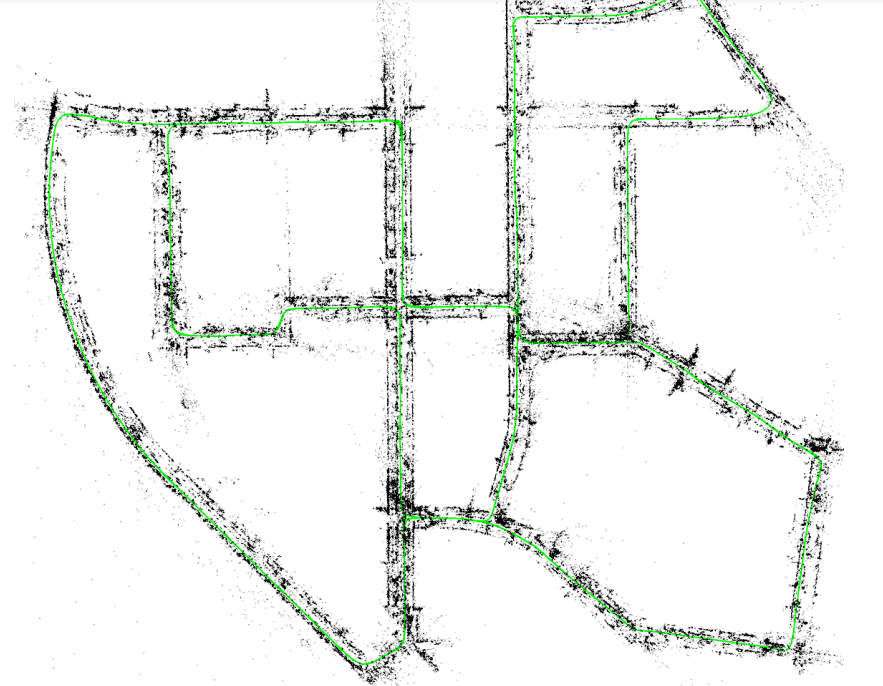}
    \caption{}
    \label{fig:orbslam}
    \end{subfigure}
    \begin{subfigure}{.4\textwidth}
    \centering
    \includegraphics[width=0.58\textwidth]{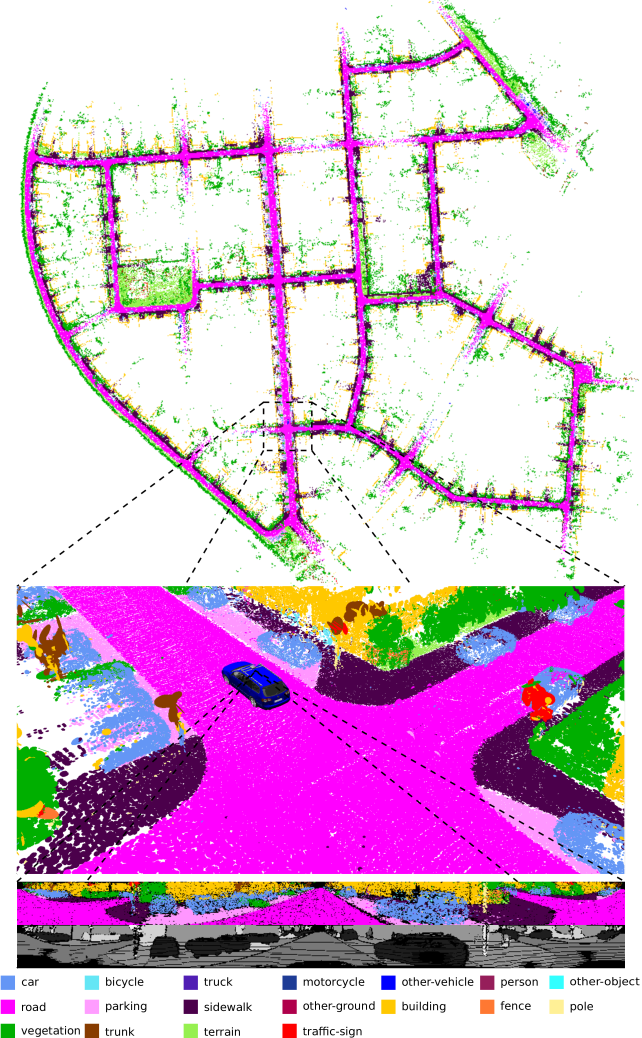}
    \caption{}
    \label{fig:suma}
    \end{subfigure}
    \label{fig:metic_semantic_map}
    \caption{(a) 3D feature map of the environment created using \textit{ORB-SLAM2}. Copyright to~\cite{orbslam2}. (b) The same environment is represented with a 3D semantic map using \textit{SUMA++}, providing richer information to understand the environment around the robot better. Copyright to~\cite{suma_plus_plus}.}
\end{figure*}

\subsubsection{Metric Factor Graphs}

Modern \ac{SLAM}, as described in~\cite{past_present_future_slam}, has moved to a more robust and intuitive representation of the state of the robot along with the sensor measurements as well as the environment map to create factor graphs as presented in~\cite{graphical_slam_for_outdoors,fastIA_olsen,graph_slam_thrun,square_root_sam,isam2}.
Factor graph based \ac{SLAM}, based on the type of map used for the environmental representation and optimization, can be divided into the \textit{Metric} and \textit{Metric-Semantic}.

\begin{table*}[!h]
    \centering
    \aboverulesep = 0pt
    \belowrulesep = 0pt
    \renewcommand{\arraystretch}{2.0}
    \scriptsize
    \setlength{\tabcolsep}{2.0pt}
    \caption{Summary of the significant VSLAM validated over public datasets.}.
    \begin{tabular}{>{\raggedright\arraybackslash}p{0.13\linewidth} | >{\raggedright\arraybackslash}p{0.13\linewidth} | >{\raggedright\arraybackslash}p{0.16\linewidth} | >{\raggedright\arraybackslash}p{0.2\linewidth} | >{\raggedright\arraybackslash}p{0.32\linewidth} }
         \toprule
         Classification & Sensors & Method & Dataset & Limitations \\
         \midrule
         \multirow{16}*{\makecell[l]{Metric \\ Factor Graphs}} & \begin{soloitemize}
             \item RGB (Mono) 
         \end{soloitemize} & ORB-SLAM \cite{orbslam} & The New College Dataset \cite{NewCollege_dataset} &  \begin{soloitemize}
             \item Difficulty in Estimating Pure Rotations
             \item Higher Error in Low Texture Environment
             \item Suffer from Motion Bias
             \item Scale Uncertainty
         \end{soloitemize} \\ \cmidrule{3-5} 
         & & DSO \cite{vi-dso}, LDSO \cite{ldso} & TUM RGB-D \cite{TUM-RGBD_dataset}, TUM-Mono \cite{TUM-Mono_Dataset}, EuRoC Mav \cite{euroc_mav_dataset}, Kitti Odometry \cite{kitti_dataset} & \begin{soloitemize}
             \item Require Constant Illumination
             \item Require Photometric Calibration for Improved Results 
             \item Scale Uncertainty
         \end{soloitemize} \\ \cmidrule{3-4}
         & & LSD-SLAM \cite{lsd_slam}, DPPTAM \cite{dpptam} & TUM RGB-D \cite{TUM-RGBD_dataset} \\ \cmidrule{3-4} 
         & & DSM \cite{dsm} & EuRoC Mav \cite{euroc_mav_dataset} \\ \cmidrule{3-5}
        & & Semi-Direct VO \cite{loosely_coupled_mono_slam} & EuRoC Mav \cite{euroc_mav_dataset}, TUM Mono \cite{TUM-Mono_Dataset} &  \begin{soloitemize}
            \item Require Accurate Initialization
            \item Scale Uncertainty
        \end{soloitemize} \\ \cmidrule{3-5}
        & & MagicVO \cite{magicvo}, DeepVO \cite{deepvo} & Kitti Odometry \cite{kitti_dataset} & \begin{soloitemize}
            \item Higher Computational Resources
            \item Less Accurate than the Classical VO Counterparts
        \end{soloitemize}\\ \cmidrule{2-5}
         & \begin{soloitemize}
             \item RGB (Mono)
             \item RGB (Stereo)
             \item RGB-D
         \end{soloitemize} & ORB-SLAM2 \cite{orbslam2} & TUM RGB-D \cite{TUM-RGBD_dataset}, EuRoC Mav \cite{euroc_mav_dataset}, Kitti Odometry \cite{kitti_dataset} & \begin{soloitemize}
             \item Lower Accuracy in High-Speed Motions
             \item Lower Accuracy in Low-Feature, Low-Lighting Environments
             \item Erroneous Loop Closures in Low-Feature/Similar Environments
         \end{soloitemize} \\ \cmidrule{2-5}
         & \begin{soloitemize}
             \item RGB (Mono)
             \item RGB (Stereo) 
             \item RGB-D
             \item RGB + IMU
         \end{soloitemize}
          & ORB-SLAM3 \cite{orbslam3} & TUM VI \cite{TUM-VI_dataset}, EuRoC Mav \cite{euroc_mav_dataset} & \begin{soloitemize}
              \item Lower Accuracy in Low-Feature, Low-Lighting Environments
              \item Erroneous Loop Closures in Low-Feature/Similar Environments
          \end{soloitemize} \\ \cmidrule{2-5}
          & \begin{soloitemize}
              \item RGB + IMU
          \end{soloitemize}  & VINS-Mono \cite{vins-fusion} &  EuRoC Mav \cite{euroc_mav_dataset} & \begin{soloitemize}
              \item Requires Good Initial Estimate
              \item Lower Accuracy in Low-Feature, Low-Lighting Environments
          \end{soloitemize}  \\ \cmidrule{3-5}
           & & SVO-Multi \cite{svo_multi} &  EuRoC Mav \cite{euroc_mav_dataset}, TUM RGB-D \cite{TUM-RGBD_dataset}, ICL-NUIM \cite{ICL_NUIM_dataset} & \begin{soloitemize}
               \item Requires Robust Initialization
               \item Flat Ground Plane Assumption causing Inaccuracies over Non-Planar Surfaces
           \end{soloitemize} \\ \cmidrule{2-5}
         & \begin{soloitemize}
             \item RGB-D
         \end{soloitemize} & CPA-SLAM \cite{ma2016cpa} & TUM RGB-D \cite{TUM-RGBD_dataset}, ICL-NUIM \cite{ICL_NUIM_dataset} & \begin{soloitemize}
             \item Lower Accuracy in case of Inaccurate Planar Detections 
         \end{soloitemize} \\ \bottomrule
         \multirow{8}*{\makecell[l]{Metric-Semantic \\ Factor Graphs}} & \begin{soloitemize}
             \item RGB (Mono)
         \end{soloitemize}  & Monocular Object SLAM \cite{real_time_obj_slam} & TUM RGB-D \cite{TUM-RGBD_dataset} & \begin{soloitemize}
             \item Pre-Generated Object Database
         \end{soloitemize} \\ \cmidrule{3-5}
         &  & QuadricSLAM \cite{quadric_slam} & TUM RGB-D \cite{TUM-RGBD_dataset} & \begin{soloitemize}
             \item Assumption of Scene Representation as Quadrics
             \item Quadric computation Computationally Expensive
             \item Scale Uncertainty
         \end{soloitemize} \\ \cmidrule{3-5}
         &  & CubeSLAM \cite{cube_slam} & TUM RGB-D \cite{TUM-RGBD_dataset}, ICL-NUIM \cite{ICL_NUIM_dataset} & \begin{soloitemize}
             \item Lower Accuracy in case of Higher Errors in Cuboid Detections 
             \item Scale Uncertainty
         \end{soloitemize} \\ \cmidrule{2-5}
         & \begin{soloitemize}
             \item RGB (Mono)
             \item RGB (Stereo)
             \item RGB-D
         \end{soloitemize}  & DynaSLAM \cite{real_time_obj_slam} & TUM RGB-D \cite{TUM-RGBD_dataset}, Kitti Odometry \cite{kitti_dataset} & \begin{soloitemize}
             \item Filter out useful Dynamic Keypoints
             \item No Topological Relationships between the Dynamic-Static Entities
         \end{soloitemize} \\ \cmidrule{2-5}
         & \begin{soloitemize}
             \item RGB-D
         \end{soloitemize}  & VDO-SLAM \cite{vdo_slam} &  Kitti Odometry \cite{kitti_dataset}, Oxford Multimotion \cite{oxford_multimotion_dataset} & \begin{soloitemize}
             \item No Topological Relations between the Dynamic-Static Entities in the Optimization Graph
         \end{soloitemize}\\ \cmidrule{2-5}
         & \begin{soloitemize}
             \item RGB (Stereo) + IMU
         \end{soloitemize}  & Kimera \cite{kimera_only} &  EuRoC Mav \cite{euroc_mav_dataset} & \begin{soloitemize}
             \item Computationally Expensive Planar Mesh Generation
             \item No Topological Constraints between the Semantic Entities
         \end{soloitemize} \\ 
         \bottomrule
    \end{tabular} \label{tab:vision_accumulated_situational_comprehension}
\end{table*}

\begin{table*}[!htb]
    \centering
    \aboverulesep = 0pt
    \belowrulesep = 0pt
    \renewcommand{\arraystretch}{2.0}
    \scriptsize
    \setlength{\tabcolsep}{2.0pt}
    \caption{Summary of the significant LIDAR-based SLAM validated over public datasets.}.
    \begin{tabular}{>{\raggedright\arraybackslash}p{0.13\linewidth} | >{\raggedright\arraybackslash}p{0.15\linewidth} | >{\raggedright\arraybackslash}p{0.13\linewidth}  | >{\raggedright\arraybackslash}p{0.18\linewidth} | >{\raggedright\arraybackslash}p{0.35\linewidth}}
         \toprule
         Classification & Sensors & Method & Dataset & Limitations \\
         \midrule
         \multirow{14}*{\makecell[l]{Metric \\ Factor Graphs}} & \begin{soloitemize}
             \item LIDAR (2D) 
         \end{soloitemize} & Cartographer \cite{cartographer} & Deutsches Museum \cite{cartographer} & \begin{soloitemize}
             \item Scan Matching can Present Inaccuracies in Cluttered/Dynamic Environments
             \item Loop Closure can Present Inaccuracies in Environments with Similar Structure  
         \end{soloitemize} \\ \cmidrule{2-5}
         & \begin{soloitemize}
             \item LIDAR (3D) 
         \end{soloitemize} & LOAM \cite{loam}, FLOAM \cite{floam}  & Kitti \cite{kitti_dataset} & \begin{soloitemize}
             \item Inaccuracies in Non-Structured Environments (without Planar/Edge Features)
             \item Inaccuracies in Presence of Dynamic Objects
             \item No Explicit Appearance Based Loop Closure 
         \end{soloitemize} \\ \cmidrule{3-5}
         &  & SUMA \cite{suma}  & Kitti \cite{kitti_dataset} & \begin{soloitemize}
             \item Require 3D LIDAR Model
             \item Errors in Loop Closures in Similar Environments
             \item Validated mostly on Outdoor Urban Environments
         \end{soloitemize} \\ \cmidrule{2-5}
         & \begin{soloitemize}
             \item LIDAR (3D) + RGB (Mono)
         \end{soloitemize} & LIMO \cite{limo} & Kitti \cite{kitti_dataset} & \begin{soloitemize}
             \item Inaccuracies in Low-Texture Environments
             \item Degradation of Performance during High-Speed Motions
             \item No Loop Closure
         \end{soloitemize} \\ \cmidrule{2-5}
         & \begin{soloitemize}
             \item LIDAR (3D) + IMU + GPS
         \end{soloitemize} & HDL-SLAM \cite{hdl_graph_slam} & Kitti \cite{kitti_dataset} & \begin{soloitemize}
             \item Scan Matching can Present Inaccuracies in Cluttered/Dynamic Environments
             \item Optimization Graph contains only Robot poses and no Environmental Landmarks
             \item Inaccurate Loop Closure in Similar Structured Environments
         \end{soloitemize} \\ 
         \bottomrule
         \multirow{6}*{\makecell[l]{Metric-Semantic \\ Factor Graphs}} & 
         \begin{soloitemize}
             \item LIDAR (3D) 
         \end{soloitemize} & LeGO-LOAM \cite{lego_loam} & Kitti \cite{kitti_dataset} & \begin{soloitemize}
             \item High Dependence on Ground Plane 
             \item Inaccuracies in Presence of Features Extracted from Dynamic Objects
         \end{soloitemize} \\  \cmidrule{3-5}
         &  & SA-LOAM \cite{cartographer} & Kitti \cite{kitti_dataset}, Semantic-Kitti \cite{semantic_kitti_dataset}, Ford Campus \cite{ford_campus_dataset} & \begin{soloitemize}
             \item Limited Accuracy in Indoor Environments
             \item Degradation in Loop Closure in Case of Noisy Semantic Detections
         \end{soloitemize}  \\ \cmidrule{3-5}
         &  & SUMA++ \cite{suma_plus_plus} & Kitti \cite{kitti_dataset}, Semantic-Kitti \cite{semantic_kitti_dataset} & \begin{soloitemize}
             \item Limited to Ourdoor Urban Environments
             \item Rely on Accurate LIDAR Model 
         \end{soloitemize} \\ 
         \bottomrule
    \end{tabular} \label{tab:lidar_accumulated_situational_comprehension}
\end{table*}

\begin{table*}[]
    \centering
    \aboverulesep = 0pt
    \belowrulesep = 0pt
    \renewcommand{\arraystretch}{2.0}
    \scriptsize
    \caption{Summary of the significant types of mapping algorithms and their limitations.}.
    \begin{tabular}{>{\raggedright\arraybackslash}p{0.15\linewidth} | >{\raggedright\arraybackslash}p{0.12\linewidth}  | >{\raggedright\arraybackslash}p{0.2\linewidth} | >{\raggedright\arraybackslash}p{0.41\linewidth}}
         \toprule
         Mapping Type & Sensors & Methods &  Limitations \\
         \midrule
         \multirow{3}*{\makecell[l]{Occupancy Maps}} & 
         \begin{soloitemize}
             \item RGB-D 
             \item 2D LIDAR
             \item 3D LIDAR
         \end{soloitemize} & Octomap \cite{hornung2013octomap} & \begin{soloitemize}
             \item Cannot Represent Exact Shape and Orientation of Objects
             \item Increased Complexity in Map Querying with Increase Map Size
             \item No Semantics
         \end{soloitemize} \\ \midrule
         \multirow{7}*{\makecell[l]{ESDF and TSDF}} &
          \begin{soloitemize}
             \item RGB-D 
             \item 3D LIDAR
         \end{soloitemize}& Voxblox \cite{Oleynikova2017} &  \begin{soloitemize}
             \item No Semantics
             \item ESDF Map Updates can Present Errors during Loop Closures 
         \end{soloitemize} \\ \cmidrule{3-4}
         & & Voxgraph \cite{reijgwart2020voxgraph} &  \begin{soloitemize}
             \item No Semantics
             \item Degradation of the ESDF Map Quality in the Presence of Noisy Odometry Estimates 
         \end{soloitemize} \\ \cmidrule{3-4}
        & & Voxblox++ \cite{voxblox_plus} & \begin{soloitemize}
             \item Degradation of the ESDF Map Quality in the Presence of Noisy Odometry Estimates 
             \item Computationally Expensive Semantic Detection 
             \item Degradation in Map Quality with Noise in Semantic Detections
         \end{soloitemize} \\
         \midrule
         \multirow{2}*{\makecell[l]{NeRF}} &
         \begin{soloitemize}
             \item RGB-D 
         \end{soloitemize} & iMap~\cite{sucar_imap_2021}, Urban Radiance Fields~\cite{rematas2022urf}, Mega-NeRF~\cite{turki2022mega}&  \begin{soloitemize}
             \item No Semantics (but potentially learnable)
             \item Computationally expensive
             \item Need to handle catastrophic forgetting while integrating new knowledge
         \end{soloitemize} \\ \midrule  
         \multirow{2}*{\makecell[l]{Surfel maps}} & \begin{soloitemize}
             \item RGB-D
             \item 3D Lidar 
         \end{soloitemize} & ElasticFusion~\cite{whelan2015elasticfusion},  SurfelMeshing~\cite{schops2019surfelmeshing},  Other~\cite{wang2019real} & \begin{soloitemize}
             \item Sparse Representation
             \item Cannot Represent Continuous Surfaces
             \item Less Useful For Path Planning and Obstacle Avoidance
         \end{soloitemize} \\
         \midrule
          \multirow{2}*{\makecell[l]{3D Scene Graphs}} & \begin{soloitemize}
             \item RGB-D 
         \end{soloitemize} & 3D DSG~\cite{3d_scene_graph}, Hydra~\cite{Hughes2022} & \begin{soloitemize}
             \item Validated mainly in Indoor Scenarios
             \item Handle Few Dynamic Objects in the Scene like Humans 
         \end{soloitemize} \\ \bottomrule
\end{tabular} \label{tab:mapping_accumulated_situational_comprehension}
\end{table*}

A metric map encodes the understanding of the scene at a geometric level (\eg lines, points, and planes), which is utilized by a \ac{SLAM} algorithm to model the environment.
\textit{Parallel Tracking and Mapping (PTAM)} is one of the first feature-based monocular algorithms which split the tracking of the camera in one thread and the mapping of the key points in another, performing batch optimization for optimizing both the camera trajectory and the mapped 3D points.
Similar extensions to the \textit{PTAM} framework are \textit{ORB-SLAM}~\cite{orbslam}, and \textit{REMODE}~\cite{remode} creating a semi-dense 3D geometric map of the environment while estimating the camera trajectory.
As an alternative to feature-based methods, direct methods use the image intensity values instead of extracting features to track the camera trajectory even in feature-less environments such as semi-dense direct \ac{VO} called \textit{DSO}~\cite{semi_dense_vo} and \textit{LDSO}~\cite{ldso} improving the \textit{DSO} by adding loop closure into the optimization pipeline, whereas \textit{LSD-SLAM}~\cite{lsd_slam}, \textit{DPPTAM}~\cite{dpptam}, \textit{DSM}~\cite{dsm}, perform a direct monocular \ac{SLAM} tracking camera trajectory along with building a semi-dense model of the environment.
Methods have also been presented that combine the advantages of both feature-based and intensity-based methods, such as \textit{SVO}~\cite{svo} performing high-speed semi-direct \textit{\ac{VO}}, \textit{CPA-SLAM}~\cite{ma2016cpa}, and \textit{Loosely coupled Semi-Direct \ac{SLAM}}~\cite{loosely_coupled_mono_slam} utilizing image intensity values for optimizing the local structure and image features for optimizing the keyframe poses.

\acl{DL} models may be used effectively to learn from data to estimate the motion from sequential observations. Hence, their online prediction could be better prior to initialize the factor graph optimization problem closer to the correct solution~\cite{yang2020d3vo, carlone2015initialization}. \textit{MagicVO}~\cite{magicvo}, and \textit{DeepVO}~\cite{deepvo} study supervised end-to-end pipelines for learning monocular \textit{\ac{VO}} from data not requiring complex formulations and calculation for several stages such as feature extraction, and matching, keeping the \ac{VO} implementation concise and intuitive.
There are also some supervised approaches like \textit{LIFT-SLAM} \cite{bruno2021lift}, \textit{RWT-SLAM} \cite{peng2022rwt}, and \cite{naveed2022deep, sun2022multi} that utilized deep neural networks for improved feature/descriptor extraction.
Alternatively, unsupervised approaches~\cite{Godard2019b, Zhou2017b, Li2018c} exploit the brightness constancy assumption between frames in close temporal proximity to derive a self-supervised photometric loss.
The methods have gained momentum, enabling learning from unlabeled videos and continuously adapting the \ac{DL} models to newly seen data~\cite{Vodisch2022a,Zhang2021c}.
Nevertheless, monocular visual-only methods suffer from a considerable limitation of being unable to estimate the metric scale directly and accurately track the robot poses in the presence of pure rotational or rapid/acrobatic motion.
\textit{RAUM-VO}~\cite{s22072651} mitigates the rotational drift by integrating an unsupervised learned pose with the motion estimated with a frame-to-frame epipolar method~\cite{kneip2013direct}.

To overcome these limitations, cameras are combined with other sensors, for example, synchronizing them with an \ac{IMU}, giving rise to the research line working on monocular \textit{\ac{VIO}}.
Methods such as \textit{OKVIS}~\cite{okvis}, \textit{SVO-Multi}~\cite{svo_multi}, \textit{VINS-mono}~\cite{vinsmono}, \textit{SVO+GTSAM}~\cite{svo_gtsam}, \textit{VI-DSO}~\cite{vi-dso}, \textit{BASALT}~\cite{basalt} are among the most outstanding examples.
\cite{vo_benchmark} benchmark all the open-source \textit{\ac{VIO}} algorithms and compare their performance on computationally demanding embedded systems.
Furthermore, \textit{VINS-fusion}~\cite{vins-fusion}, \textit{ORB-SLAM2}~\cite{orbslam2} (see Fig.~\ref{fig:orbslam}) provide a complete framework capable of fusing either \textit{monocular}, \textit{stereo} or \textit{RGB-D} cameras with an \ac{IMU} to improve the overall tracking accuracy of the algorithms.
\textit{ORB-SLAM3}~\cite{orbslam3} presents improvement over \textit{ORB-SLAM2} performing even multi-map \ac{SLAM} using different visual sensors along with an \ac{IMU}.

Methods have been presented that perform thermal inertial odometry for performing autonomous missions using robots in visually challenging environments, such as~\cite{thermal_IO,graph_based_planner,autonomous_search,tp-tio}.
Authors in \textit{TI-SLAM}~\cite{ti-slam} not only perform thermal inertial odometry but also provide a complete \ac{SLAM} backend with thermal descriptors for loop closure detections.
\cite{event_vio} presents a continuous-time integration of event cameras with \ac{IMU} measurements, improving by almost a factor of four the accuracy over event only \textit{EVO}~\cite{evo}.
Ultimate \ac{SLAM}~\cite{ultimate_slam} combines RGB cameras with event cameras along with \ac{IMU} information to provide a robust \ac{SLAM} system in high-speed camera motions.

\ac{LIDAR} odometry and \ac{SLAM} for creating metric maps have been widely researched in robotics to create metric maps of the environment such as \textit{Cartographer}~\cite{cartographer}, 
\textit{Hector-SLAM}~\cite{hector_slam} performing a complete \ac{SLAM} using 2D \ac{LIDAR} measurements and \textit{LOAM}~\cite{loam} providing a \textit{parallel \ac{LIDAR} odometry and mapping} technique to simultaneously compute the \ac{LIDAR} velocity while creating accurate 3D maps of the environment.
To further improve the accuracy, techniques have been presented which combine vision and \ac{LIDAR} measurement as in \textit{Lidar-Monocular Visual Odometry (LIMO)}~\cite{limo}, \textit{LVI-SLAM}~\cite{lvi_slam} combining robust monocular image tracking with precise depth estimates from \ac{LIDAR} measurements for motion estimation.
Methods like \textit{LIRO}~\cite{liro}, \textit{VIRAL-SLAM}~\cite{viral_slam}, a couple of additional measurements like \textit{Ultra Wide Band (UWB)} with visual and \ac{IMU} sensors for robust pose estimation and map building.
Other methods like \textit{HDL-SLAM}\cite{hdl_graph_slam}, \textit{LIO-SAM}~\cite{lio_sam} tightly couple along with \ac{IMU}, \ac{LIDAR} and \ac{GPS} measurements, for globally consistent maps. 

While significant progress has been demonstrated using \textit{metric \ac{SLAM}} techniques, one of the major limitations of these methods is the lack of information extracted from the metric representation, such as (1) \textit{No semantic knowledge of the environment}, (2) \textit{Inefficiency in identifying static and moving objects}, and (3) \textit{Inefficiency in distinguishing different object instances}.

\subsubsection{Metric-Semantic Factor Graphs}
As explained in Sect.~\ref{subsec:scene_understanding}, the advancements in \textit{direct situational comprehension} techniques have enabled a higher-level understanding of the environments around the robot, leading to the evolution of \textit{metric-semantic \ac{SLAM}} overcoming the limitations of traditional \textit{metric \ac{SLAM}} enabling the robot with the capabilities of human-level reasoning.
Several approaches to address these solutions have been explored, which will be discussed here.

\textbf{Object-based Metric-Semantic \ac{SLAM}.} build a map of the instances of the different detected object classes on the given input measurements.
The pioneer works \textit{SLAM++}~\cite{slam++} and~\cite{real_time_obj_slam} create a graph using camera pose measurements and the objects detected from previously stored database to optimize the camera and the object jointly poses.
Following these methods, many object-based \textit{metric-semantic \ac{SLAM}} techniques have been presented, such as~\cite{localization_using_permanent_matrix, probablisitic_bowman,vso,quadric_slam,cube_slam,probablistic_doherty,vps_slam,semantic_jose} not requiring a previously stored database and jointly optimizing the camera poses, 3D geometric landmarks as well as the semantic object landmarks.
\textit{SA-LOAM}~\cite{sa_loam} utilize semantically segmented 3D \ac{LIDAR} measurements for generating a semantic graph for robust loop closures. 
The primary sources of inaccuracies of these techniques are due to extreme dependence on the existence of objects, as well as (1) \textit{uncertainty in object detection}, (2) \textit{partial views of the objects which are still not handled efficiently} (3) \textit{no consideration of the topological relationship between the objects}.
Moreover, most of the previously presented approaches cannot handle dynamic objects.
Research works on adding dynamic objects to the graph such as \textit{VDO-SLAM}~\cite{vdo_slam}, \textit{RDMO-SLAM}~\cite{rdmo_slam} reduce the influence of the dynamic objects on the optimized graph.
Nevertheless, they can not handle complex dynamic environments and only generate a sparse map without topological relationships between these dynamic elements.

\textbf{SLAM with Metric-Semantic Map.} augments the output metric map given by \ac{SLAM} algorithms with semantic information provided by \textit{scene understanding} algorithms, as~\cite{semantic_rtab_map, Lai2020, Hempel2022},  or \textit{SemanticFusion}~\cite{semantic_fusion} and \textit{Kimera-Multi}~\cite{Tian2022}.
These methods assume a static environment around the robot; thus, the quality of the \textit{metric-semantic map} of the environment can degrade in the presence of moving objects in the background.
Another limitation of these methods is that they do not utilize useful semantic information from the environment to improve the robot's pose estimation, thus, the map quality.

\textbf{SLAM with Semantics to Filter Dynamic Objects.} utilizes the available semantic information of the input images provided by the \textit{scene understanding} module, only to filter bad-conditioned objects (\ie moving objects) from pictures given to the \ac{SLAM} algorithms, as~\cite{computational_eff_semantic_slam, liu2019dms, li2021dp} for image-based, or \textit{SUMA++}~\cite{suma_plus_plus} (see Fig.~\ref{fig:suma}) for \ac{LIDAR} based. Although these methods increase the accuracy of the \ac{SLAM} system by filtering moving objects, they neglect the rest of the semantic information from the environment to improve the robot's pose estimation.
Table.~\ref{tab:vision_accumulated_situational_comprehension} and Table.~\ref{tab:lidar_accumulated_situational_comprehension} provide a brief summary of the above-presented approaches highlighting the different datasets used in their validation along with their limitations. 

\subsection{Mapping}
\label{subsec:scene_graphs}

This section covers the recent works which focus only on the complex high-level representations of the environment.
Most of these methods assume the \ac{SLAM} problem to be solved and focus only on the scene representation.
An ideal environmental representation must be efficient concerning the required resources, capable of reasonably estimating regions not directly observed, and flexible enough to perform reasonably well in new environments without any significant adaptations.

Occupancy mapping is a method for constructing an environment map in robotics. 
It involves dividing the environment into a grid of cells, each representing a small portion of the environment. 
The occupancy of a cell represents the likelihood of that cell being occupied by an obstacle or not. 
Initially, all cells in the map can be considered unknown or unoccupied.
As the robot moves and senses the environment, the occupancy of cells is updated based on sensor data. 
One of the most popular approaches in this category is Octomap~\cite{hornung2013octomap}. 
They represent the grid of cells through a hierarchical structure that allows a more efficient query of the occupancy probability in a certain space.

The adoption of \ac{SDF}-based in robotics is well established to represent the robot's surroundings~\cite{oleynikova2016signed} to enable planning a trajectory towards the mission goal safely~\cite{oleynikova2018safe}. In general, \ac{SDF} is a three-dimensional function that maps points of a metric space to the distance to the nearest surface. \ac{SDF} can represent distances in any number of dimensions, including two-dimensional and higher-dimensional spaces, and can represent complex geometries and shapes with an arbitrary curvature and are therefore widely used in computer graphics. However, a severe limitation of \textit{SDF} is that they can only represent watertight surfaces, \ie surfaces that divide the space into inside and outside~\cite{chibane_neural_2020}.

\ac{SDF} has two main variations, \ac{ESDF} and \ac{TSDF}, which usually apply to a discretized space made of voxels. 
On the one hand, \ac{ESDF} gives the distance to the closest obstacle for free voxels and the contrary for occupied ones.
They have been used for mapping in \textit{FIESTA}~\cite{Han2019}, where the authors exploit the property of direct modeling free space for collision checking and the gradient information for planning~\cite{zucker2013chomp} while dramatically improving their efficiency.
On the other hand, \ac{TSDF} relies on projective distance, which is the distance measured along the sensor ray from the camera to the observed surface. 
The distances are calculated only within a specific radius around the surface boundary, known as the truncation radius~\cite{Oleynikova2017}. 
This helps improve computational efficiency and reduce storage requirements while accurately reconstructing the observed scene.
\ac{TSDF} has been demonstrated in multiple works such as \textit{Voxgraph}~\cite{reijgwart2020voxgraph}, \textit{Freetures}~\cite{millane2020freetures}, \textit{Voxblox++}~\cite{voxblox_plus}, or the more recent \textit{Voxblox-Field}~\cite{Pan2022}.
They can create and maintain globally consistent volumetric maps that are lightweight enough to run on computationally constrained platforms and demonstrate that the resulting representation can navigate unknown environments. 
Panoptic segmentation has been integrated with \ac{TSDF} by \cite{naritaPanopticFusionOnlineVolumetric2019} for labeling each voxel semantically while differentiating between \textit{stuff}, \eg the background wall and floor, from \textit{things}, \eg movable objects. Furthermore, \cite{schmidPanopticMultiTSDFsFlexible2022} leverage pixel-wise semantics to maintain temporal consistency and detect changes in the map caused by movable objects, hence surpassing the limitations of a static environment assumption.

\textit{Implicit Neural Representations (INR)} (sometimes also referred to as coordinate-based representations) are a novel way to parameterize signals of all kinds, even environments parameterized as 3D points clouds, voxels, or meshes.
With this in mind, \textit{Scene Representation Networks (SRNs)}~\cite{sitzmann2019srns} are proposed as a continuous scene representation that encodes both geometry and appearance and can be trained without any 3D supervision.
It is shown that \textit{SRNs} generalize well across scenes, can learn geometry and appearance priors, and are helpful for novel view synthesis, few-shot reconstruction, joint shape, and appearance interpolation unsupervised discovery of non-rigid models.
In~\cite{sitzmann_implicit_2020}, a new approach is presented, capable of modeling signals with fine details and accurately capturing their spatial and temporal derivatives.
Based on periodic activation functions, this approach demonstrates that the resulting neural networks referred to as \textit{Sinusoidal Representation Networks (SIRENs)} are well suited for representing complex signals, including 3D scenes. 

\textit{Neural Radiance Fields (NERF)}~\cite{mildenhall2020nerf} exploits the framework of \textit{INR}s to render realistic 3d scenes by a differential process that takes as input a ray direction and predicts the color and density of the scene along that ray.
\cite{sucar_imap_2021} pioneered the first application of \textit{NERF} to \ac{SLAM} for representing the knowledge of the 3D structure inside the weights of a deep \ac{NN}. For their promising results, this research prospect attracted numerous following works that continuously improve the fidelity of the reconstructions and the possibility of updating the knowledge of the scene while maintaining previously-stored information~\cite{zhu2023nicer, Zhu2022CVPR, rosinol2022nerf}.

Differently from the previous dense environment representation methods, which are helpful for autonomous navigation, sparser scene representation also exists, such as point clouds and surfel maps, which are more commonly used for more straightforward tasks such as localization. Remarkably, a surfel, \ie a surface element, is defined by its position in 3D space, the surface normal, and other attributes such as color and texture. Their use has been extensively explored in recent \ac{LIDAR}-based \ac{SLAM} to efficiently represent a 3D map that can be performed as a consequence of optimization following revisited places, \ie loop closure~\cite{schops2019surfelmeshing, wang2019real}.

\begin{figure*}[t]
    \centering
    \includegraphics[width=0.5\textwidth]{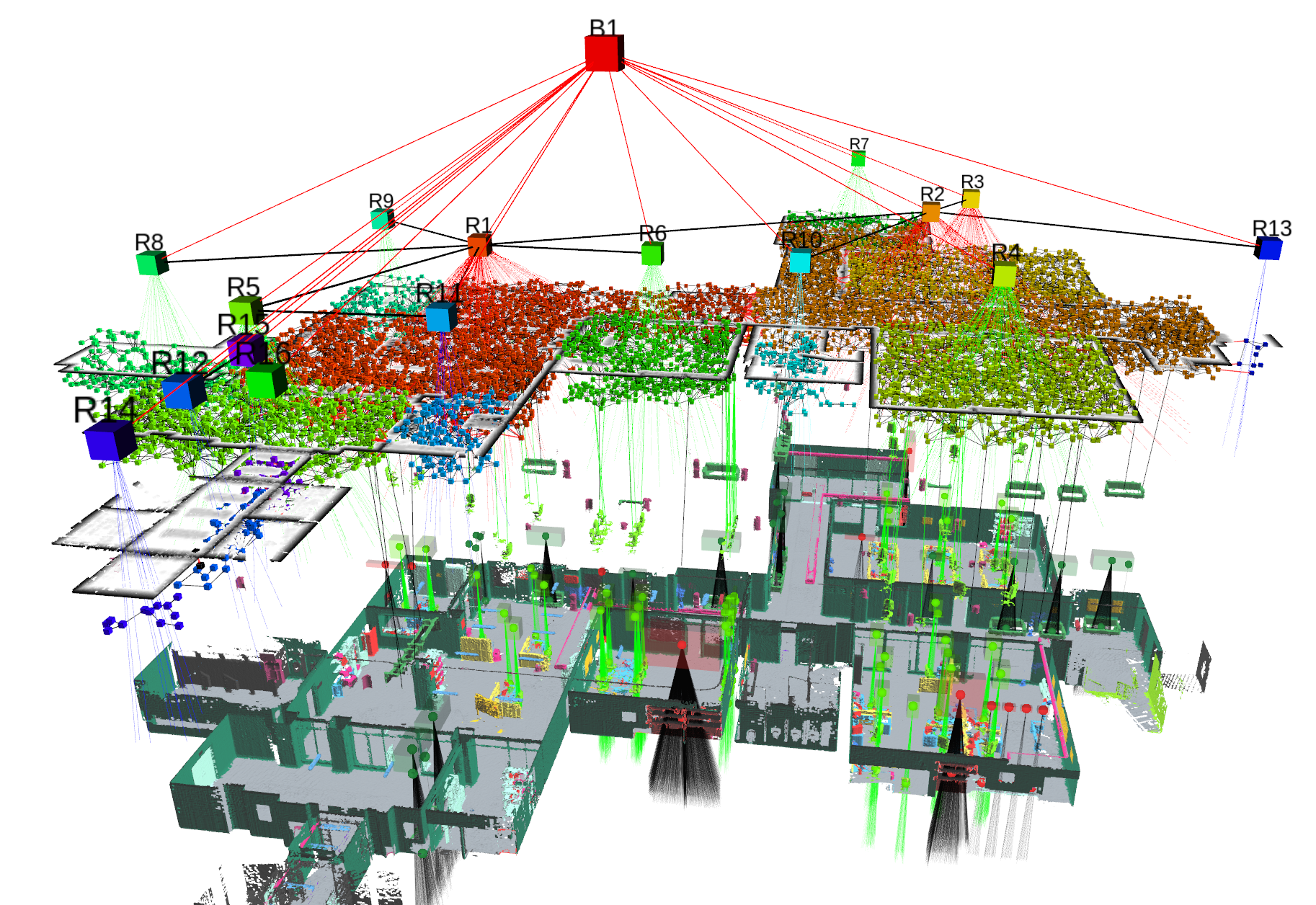}
    \caption{A \acl{DSG} generated by~\cite{kimera} with a multi-layer abstraction of the environment. Copyright to~\cite{kimera}.}
    \label{fig:dsg}
\end{figure*}
3D \textit{scene graphs} have also been researched to represent a scene, such as~\cite{3d_scene_graph, learning_3d_semantic_scene_graphs, Wu2021}, which build a model of the environment, including not only metric and semantic information but also essential topological relationships between the objects of the environment.
They can construct an environmental graph spanning an entire building, including the semantics of objects (class, material, and shape), rooms, and the topological relationships between these entities.
However, these methods are executed offline and require an available 3D mesh of the building with the registered RGB images to generate the \textit{3D scene graphs}.
Consequently, they can only work in static environments.

\acp{DSG} (see Fig.~\ref{fig:dsg}) are an extension of the aforementioned \textit{scene graphs} to include dynamic elements (\eg humans) of the environment in an actionable representation of the scene that captures geometry and semantics~\cite{Rosinol2020}.
\cite{kimera} present the first method to build a \ac{DSG} automatically using the input of a \ac{VIO}~\cite{kimera_only}. Furthermore, it allows tracking of the pose of humans and optimizes the mesh based on the deformation of the space induced by detected loop closures.
Although these results are promising, their main drawback is that the \ac{DSG} is built offline, and the \ac{VIO} first creates a 3D mesh-based semantic map fed to the dynamic scene generator.
Consequently, the \ac{SLAM} does not use these topological relationships to improve the accuracy of the spatial reconstruction of the robot trajectory. Besides, except for the humans, the rest of the topological relationships are considered purely static, \eg chairs or other furniture are fixed to the first detection location.

More recently, Hydra~\cite{Hughes2022} implements the scene graph construction into a real-time capable system relying on a highly parallelized architecture. 
Moreover, they can optimize an embedded deformation graph online, consequently to a loop closure detection. 
Remarkably, the information in the graph allows the creation of descriptors based on objects and visited places histograms that can be matched robustly with previously seen locations.
Therefore, \acp{DSG}, even if in their infancy stage, are shown to be a practical decision-making tool that enables robots to perform autonomous tasks. 
For example,~\cite{Ravichandran2022} demonstrate how they can be used for learning a trajectory policy by turning a \ac{DSG} into a graph observation that serves as input to a \ac{GNN}.
Or \ac{DSG} may be used for planning challenging robotic tasks as proposed in the Taskography benchmark~\cite{agia2022taskography}. 

Lastly, one of the main features of a \ac{DSG} is the possibility to perform queries and predictions of the future evolution of the scene based on dynamic models linked with the agents or physical elements~\cite{Rosinol2020}. And an even more intriguing property is their application to scene change detection or long-term the newly formalized \textit{semantic scene variability estimation} task, which sets as a goal the prediction of long-term variation in location, semantic attributes, and topology of the scene objects~\cite{Looper2022}.
This property has only been explored by scratching the surface of its potential application.
Still, it already grasps our vision of a comprehension layer that produces the knowledge required by the projection and prediction of future states.
Table.~\ref{tab:mapping_accumulated_situational_comprehension} summarizes the main mapping methods described above with their generated map types and limitations. 
\section{Situational projection}
\label{sec:projection}

In robotics, the projection of the situation is essential for reasoning and execution of a planned mission~\cite{motion_planning_manuel}. 
However, it has primarily focused on the prediction of the future state of the robot by using a dynamic model~\cite{iekf}. 
At the same time, most of the works assume a static time-invariant environment.
Some other works~\cite{semantic_jose, kimera, multimodal_kalman_filter} have incorporated dynamic models on some environment elements, such as persons or vehicles, and, to a certain extent, the uncertainty of the motion. 

The projection component requires more effort in producing models that can forecast the dynamic agents' behavior and how the scene is affected by changes that shift its appearance over time.
Remarkably, numerous research areas address specific forecast models, the interactions between agents, and the surrounding environment's evolution.
Below, we give an overview of the most prominent that we deem more related to the robotic \ac{SA} concept.

\subsection{Behavior Intention Prediction}

Behavior Intention Prediction (BIP) focuses on developing methods and techniques to enable autonomous agents, such as robots, to predict the intentions and future behaviors of humans and other agents they interact with. This research is essential for effective communication, collaboration, and decision-making. BIP typically involves integrating information from multiple sources, such as visual cues, speech, and contextual information, \eg coming from the comprehension layer. This research has numerous applications, including human-robot collaboration, autonomous driving, and healthcare. Especially regarding the \ac{AV} application, this topic has gained importance among researchers and is widely studied due to safety concerns. However, we argue that the outcome of its investigation may apply to other tasks implying interaction among a multitude of agents. 

To define \ac{AV} BIP task, we refer to the recent survey~\cite{Fang2022} that distinguishes various research topics related to understanding the driving scene under a unified taxonomy.
The whole problem is then defined by analyzing on a timeline the events happening on the road scenario and the decision-making factors that lead to specific outcomes. 

Scene contextual factors, such as traffic rules, uncertainties, and interpretation of goals, are crucial to inferring the interaction among road actors~\cite{Rasouli2020} and the safety of the current driving policy~\cite{Guo2020}. 
Specifically, interaction may be due to social behavior or physical events such as obstacles or dynamic clues, \eg traffic lights, that influence the decision of the driver~\cite{Wang2022}.
Multi-modal perception is exploited to infer whether pedestrians are about to cross~\cite{Kwak2017} or vehicles to change lanes~\cite{Xing2019}. 
Mostly, recent solutions rely on \ac{DL} models such as \acp{CNN}~\cite{Fang2020, Izquierdo2019}, \ac{RNN}~\cite{Wang2022, Rasouli2022}, \acp{GNN}~\cite{Cadena2022} or on the transformer attention mechanism, which can estimate the crossing intention using only pedestrian bounding boxes as input features~\cite{Achaji2022}. 
Otherwise, causality relations are studied by explainable AI models to make risk assessment more intelligible~\cite{Li2020}. 
Lastly, simulation tools of road traffic and car driving, such as CARLA~\cite{dosovitskiy2017carla}, can be used as forecasting models provided that mechanisms to adapt synthetically generated data to the realistic are put in place~\cite{Zhou2022}.

BIP then requires fulfilling the task of predicting the trajectory of the agents. For an ~\ac{AV}, the input to the estimation is represented by the historical sequence of coordinates of all traffic participants, plus possibly other contextual information, \eg velocity. The task is then to generate a plausible progression of the future position of other pedestrian vehicles. Methods for predicting human motion have been exhaustively surveyed by \cite{Rudenko2020}, and regarding vehicles~\cite{Huang2022}, who classified the approaches into four main categories: physics-based, machine-learning, deep-learning, and reinforcement learning. Moreover, the authors determine the various contextual factors that may constitute additional input for the algorithms similar to those previously described. Finally, they acknowledge that complex deep learning architectures are the \textit{de-facto} solution for real-world implementation for their performance.

Additionally, \ac{DL} allows for multi-modal outputs, \ie generation of diverse trajectory with an associated probability, and for multi-task learning, \ie producing a likelihood of specific behavior simultaneously. Behavior prediction is, in fact, a separate task more concerned with assigning to the road participants an intention of performing a particular action. Reviews of approaches specific for understanding the behaviors of vehicles~\cite{Mozaffari2022} and pedestrians~\cite{Ridel2018} is found in the recent literature. Behavior prediction is also related to forecasting the occurrence of accidents. This capability is a highly demanded skill in many industrial scenarios.

\section{Discussion}
\label{sec:discussions}

In the previously presented sections, we thoroughly review the state-of-the-art techniques presented by the scientific community to improve the overall intelligence of autonomous robotic systems. 
Importing the knowledge from psychology to robotics, we show that a situational awareness perspective in robotics can embark efficiently on these presented state-of-the-art techniques in an organized and multi-layered manner. Based on our presented survey, we address the research questions posed earlier:

Through the literature review, we find a gap between the presented approaches to provide a unified and complete \acl{SA} for the robots to understand and reason about the environment so as to perform a mission autonomously closely to human beings. To this end, we proposed an ideal model of the robotic \ac{SA} system, which, per our mentioned conventions, would be divided into 3 sub-sections. The \textbf{Perception layer} should consist of a multi-modal sensor suite for accurate environmental perception. The \textbf{Comprehension layer} may bear methods from \textit{direct situational comprehension} and \textit{accumulated situational comprehension} to improve the robot's ego-awareness of its state, such as the pose, but also model the external factors with which it interacts, \eg objects and the environment 3D structure, in the form a metric-semantic-topological scene graphs. The \textbf{Projection layer}, which still has few connections with the underlying perception and comprehension and is usually treated on a standalone basis, would add forecasting models to predict the future state of the robot as well as the dynamic environmental elements.

\begin{figure}[t]
    \centering
    \includegraphics[width=0.5\textwidth]{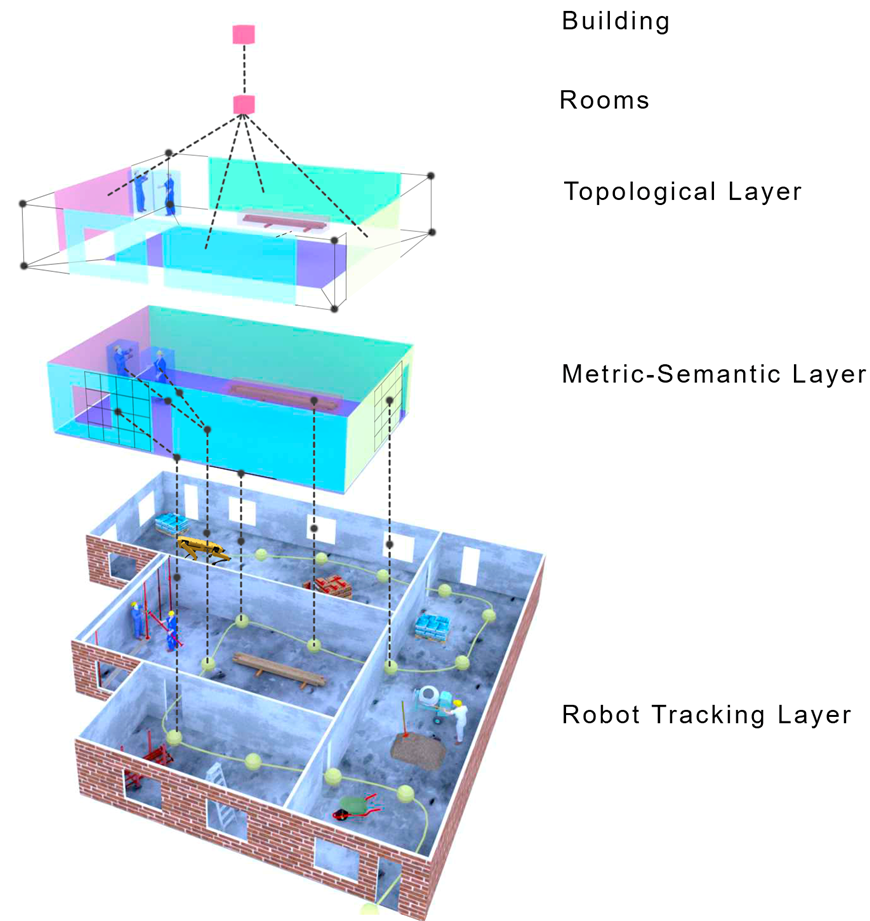}
    \caption{Proposed \textit{\ac{S-Graph}}. The graph is divided into three sub-layers: the tracking layer, which tracks the sensor measurements as it creates a local keyframe map containing its respective sensor measurements. The metric-semantic layer creates a metric-semantic map using the local keyframes. The topological layer consists of the topological connections between the elements in a given area and the rooms that connect planar features.}
    \label{fig:scene_graph}
\end{figure}

The progress in \ac{AI} and \ac{DL} has been pivotal in enhancing the robot's comprehension of the situation, as depicted by previous research. 
Despite significant progress in mobile robot \textit{direct situational comprehension}, a versatile and standardized sensor suite must be developed to handle environmental challenges, such as meteorological changes.
Additionally, integrating these algorithms efficiently into scene modeling frameworks remains a challenge. 

The scene graph presented in the related works is a widely adapted term  in computer vision~\cite{Chang2023} to describe the relationship between objects in the scene with a structured representation between entities and predicates usually built from visual information. 
However, we see the current scene graphs used in mobile robotics are insufficient to address complex autonomous tasks, such as multi-modal open-set queryable maps for navigation~\cite{Huang2023, Jatavallabhula2023}. Therefore, starting from the scene graph concept, we introduce the \textit{\ac{S-Graph}} as a knowledge graph that emphasizes the ability to store the entire representation of the situation, comprising the currently perceived aspects of the scene, their comprehension, integration with previous records or possibly also external sources from a standardized ontology~\cite{CornejoLupa2021}, and the prediction of the future by the projection of the entities through their models.

Hence we set \textit{\ac{S-Graph}} (see Fig.~\ref{fig:scene_graph}) as a future target of the evolution of the current scene graphs, which add a hierarchy of conceptual layers that contribute to including prior knowledge of the situation while maintaining their formulation.
Furthermore, the current implementation of \textit{\ac{S-Graph}}~\cite{s_graphs, s_graphs+}, which is still in its initial stage, stresses the practical characteristic of using the created entity relationships, \eg topological, for obtaining an optimized answer on the state of an autonomous agent, \eg the robot's pose.

We believe this approach can accelerate progress and improve the autonomy of mobile robots.

\section{Conclusions}
\label{sec:conclusion}

In this paper, we argued that \acl{SA} is an essential capability of humans that has been studied in several different fields but has barely been considered in robotics. 
Instead, robotic research has focused on ideas in a diversified manner, such as sensing, perception, localization, and mapping.
Thus, as a direct line of future work, we proposed a three-layered \acl{SA} framework composed of perception, comprehension, and projection.
To this end, we provided a thorough literature review of the state-of-the-art techniques for improving robotic intelligence. Then, we reorganized them in a more structured, layered perception, comprehension, and projection format. 

Finally, we conclude by providing appropriate answers to the earlier research question.  

\begin{itemize}
     \item \textit{What has been achieved so far, and what challenges remain?} \\
     Given the advancements in \ac{AI} and \ac{DL}, we notice an improved comprehension layer by evaluating state-of-the-art algorithms. Comparing the initial approaches relying on heuristics and heavily engineered processing, current algorithms can solve complex tasks requiring generalization and adaptation in dynamic environments. Nevertheless, the algorithms follow a compartmentalized approach impeding a unified SA for mobile robots. Remarkably, forecasting the future situation is also in its infancy and relies on perfect data from the perception and comprehension layers to demonstrate meaningful results.      
     
    \item \textit{What could be the future direction of \acl{SA}?} \\
    We argue that after analyses of these algorithms, a situational awareness perspective can steer for faster achievement of robots, comprising of multi-modal hierarchical \textit{S-Graphs} generating a metric-semantic-topological map of its environment as well as improving the robots pose uncertainty in it. We foresee \textit{S-Graph} will be characterized by a tighter coupling of situational projection, perception, and comprehension, to complete the transition from static world assumptions to natural dynamic environments.     
\end{itemize}

\IEEEpeerreviewmaketitle

\bibliographystyle{IEEEtran}
\bibliography{IEEEabrv,Bibliography}
\vfill
\end{document}